\documentclass{article}

\PassOptionsToPackage{numbers, compress}{natbib}

\usepackage[final]{neurips_2024}

\usepackage{graphicx}
\usepackage{multirow}
\usepackage{kotex}
\usepackage[utf8]{inputenc} 
\usepackage[T1]{fontenc}    
\usepackage{hyperref}       
\usepackage{url}            
\usepackage{booktabs}       
\usepackage{amsfonts}       
\usepackage{nicefrac}       
\usepackage{microtype}      
\usepackage{xcolor}         
\usepackage{xurl}
\usepackage{pifont}
 
\newcommand{\cmark}{\ding{51}}%
\newcommand{\xmark}{\ding{55}}%

\title{Paralinguistics-Aware Speech-Empowered Large Language Models for Natural Conversation}

\author{%
  \textbf{Heeseung Kim}$^{1}$ \quad \textbf{Soonshin Seo}$^{2}$ \quad \textbf{Kyeongseok Jeong}$^{2}$ \quad \textbf{Ohsung Kwon}$^{2}$ \quad \textbf{Soyoon Kim}$^{2}$ \\ \textbf{Jungwhan Kim}$^{2}$ \quad \textbf{Jaehong Lee}$^{2}$ \quad \textbf{Eunwoo Song}$^{2,4}$ \quad \textbf{Myungwoo Oh}$^{2}$ \quad \textbf{Jung-Woo Ha}$^{2,3}$ \\ \textbf{Sungroh Yoon}$^{1,4,5}$\thanks{Corresponding authors: Kang Min Yoo \texttt{<kangmin.yoo@navercorp.com>}, Sungroh Yoon \texttt{<sryoon@snu.ac.kr>}} \quad \textbf{Kang Min Yoo}$^{2,3,4}$\footnotemark[1] \\ \\
  $^{1}$Data Science and AI Lab, Department of ECE, Seoul National University\\
  $^{2}$NAVER Cloud \quad $^{3}$NAVER AI Lab\\
  $^{4}$Artificial Intelligence Institute, Seoul National University\\
  $^{5}$ASRI, INMC, ISRC, and Interdisciplinary Program in AI, Seoul National University\\
}

\begin{document}

\maketitle

\begin{abstract}
Recent work shows promising results in expanding the capabilities of large language models (LLM) to directly understand and synthesize speech. However, an LLM-based strategy for modeling spoken dialogs remains elusive, calling for further investigation. This paper introduces an extensive speech-text LLM framework, the Unified Spoken Dialog Model (USDM), designed to generate coherent spoken responses with naturally occurring prosodic features relevant to the given input speech without relying on explicit automatic speech recognition (ASR) or text-to-speech (TTS) systems. We have verified the inclusion of prosody in speech tokens that predominantly contain semantic information and have used this foundation to construct a prosody-infused speech-text model. Additionally, we propose a generalized speech-text pretraining scheme that enhances the capture of cross-modal semantics. To construct USDM, we fine-tune our speech-text model on spoken dialog data using a multi-step spoken dialog template that stimulates the chain-of-reasoning capabilities exhibited by the underlying LLM. Automatic and human evaluations on the DailyTalk dataset demonstrate that our approach effectively generates natural-sounding spoken responses, surpassing previous and cascaded baselines. Our code and checkpoints are available at \href{https://github.com/naver-ai/usdm}{https://github.com/naver-ai/usdm}.
\end{abstract}

\section{Introduction}

Large language models (LLMs) have gained significant traction thanks to emergent capabilities \cite{openai2023gpt4, wei2022emergent, brown2020language, kaplan2020scaling, yoo2024hyperclovaxtechnicalreport}, such as few-shot in-context learning, complex reasoning \cite{wei2022chain, yao2022react}, and instruction-following \cite{ouyang2022training}.
These remarkable discoveries led to chat-enabled LLMs and generative personal assistants \cite{gabriel2024ethics}. 
However, text-based agents are limited in usability due to their medium of interaction. 
Ideally, speech-enabled LLMs would recognize the user's emotional state or subtle nuance and generate spoken responses with prosody most appropriate to the user's context. 
Although automatic speech recognition (ASR) and text-to-speech (TTS) systems can be easily employed, the linguistic discrepancy between speech and text causes dialog inefficiencies and result in sub-optimal user experience \cite{horowitz1987comprehending, clark2019state}.
As such, systematically integrating the speech modality into LLMs can unlock speech interactivity while retaining LLMs' powerful capabilities.

\begin{figure}[t]
    \centering
    \includegraphics[width=0.57\linewidth]{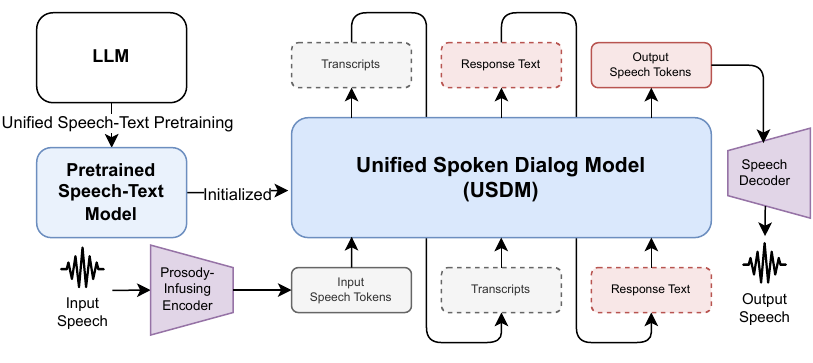} 
    \includegraphics[width=0.41\linewidth]{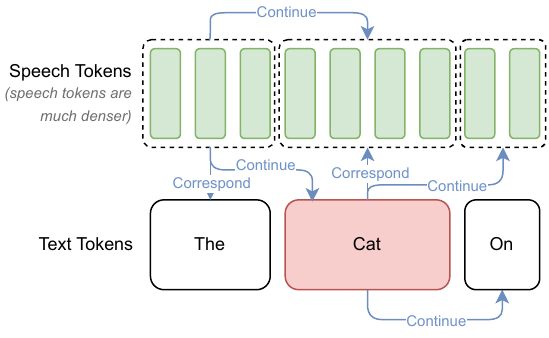} 
    \caption{Overview of our spoken dialog modeling approach (Left). All possible self-supervised learning objectives from our speech-text pretraining scheme. (Right)}
    \label{fig:usdm-overview-intro}
    \vskip -0.2in
\end{figure}

Recent advances have also spurred the idea of large foundational models (LFM) for other modalities (e.g., vision, speech, etc.) \cite{bommasani2021opportunities}, unifying LLMs with the sensory spaces. In the vision domain, numerous works have explored interfacing pretrained language models with the visual modality \cite{tsimpoukelli2021multimodal, alayrac2022flamingo, cho2021unifying, yu2023scaling}. 
More recently, \citet{yu2023scaling} proposed a vision language model (VLM) based on a pretrained LLM that directly generates discrete vision tokens, which can be decoded into high-fidelity images. 
Such work shows that autoregressive language models can model the vision modality. 

In the speech domain, while earlier work focused on text-less speech modeling \cite{lakhotia-etal-2021-generative, chen2022wavlm, nguyen-etal-2023-generative}, recent work has either taken inspiration from the LLM architecture to achieve speech synthesis \cite{wang2023neural} or incorporate pretrained language models into speech-understanding tasks \cite{chen2022wavlm, arjmand2021teasel, ao2021speecht5, chu2023qwen}, where the model is limited to text outputs.
More recent work explores the possibility of empowering pretrained LLMs to autoregressively generate discrete speech tokens for speech translation \cite{rubenstein2023audiopalm} and speech-instruction following tasks \cite{zhang2023speechgpt, yang2023uniaudio, chen2023salm}. 
Despite these successes, more work is needed to understand whether LLMs are capable of generating speech, understanding, and incorporating paralinguistics that are appropriate and natural for the social context, especially in spoken dialog settings.

We introduce the Unified Spoken Dialog Model (USDM), a novel LLM-based approach for modeling spoken dialogs in an end-to-end fashion. 
We propose a novel speech-text pretraining scheme that promotes learning cross-modal distributional semantics, which is vital for imbuing LLMs with the ability to generate coherent speeches in spoken dialog modeling.
In particular, based on the observation that any subsample of either speech or text in corresponding speech-text pairs form two types of relationships with the other modality (right side of Figure \ref{fig:usdm-overview-intro}), we formulate a large number of combinations of training objectives that theoretically benefits all speech-text tasks, including spoken dialog modeling. 
We then fine-tune our pretrained speech-text model with spoken dialog data, breaking the speech-to-speech modeling problem into intermediary steps that are easier for the underlying pretrained LLM to handle (Figure \ref{fig:usdm-overview-intro}).
To enhance the effects of our speech-text pretraining and spoken dialog modeling, in addition, we adopt the prosody-infused speech tokenization scheme based on the discovery that the speech token, previously used to convey semantic information, also contains prosody information.

We demonstrate that USDM outperforms baselines in spoken dialog modeling for the DailyTalk dataset. We further validate the effectiveness of our pretraining and fine-tuning schemes through comprehensive ablation studies. Along with various analyses, we highlight the capabilities of USDM with diverse samples on our demo page.\footnote{Demo: \href{https://unifiedsdm.github.io/}{https://unifiedsdm.github.io/}} Our contributions are as follows.

\begin{itemize}
    \item We propose a unified pretraining strategy for modeling the comprehensive relationship between the speech and text modalities that is especially effective for downstream speech-to-speech spoken dialog generation.

    \item We present an extensive spoken dialog modeling framework detailing the discrete speech tokenization scheme utilizing a pair of a prosody-infusing encoder and a decoder. Additionally, we propose an LLM-based modeling strategy for generating natural-sounding and semantically coherent dialog responses in an end-to-end fashion.

    \item Our work establishes the foundation for speech-enabled chat-based LLMs, showcasing a prototype that not only maintains the LLM's ability to generate dialog responses but also enhances LLM with speech-interaction capabilities. 
\end{itemize}

\section{Related Work}

\paragraph{Discrete Speech Representations.}
\label{background:token}
To construct spoken language models (SLM), various discrete speech representations have been utilized in previous works \cite{lakhotia-etal-2021-generative, wang2023neural, 10.1162/tacl_a_00618, borsos2023audiolm,hassid2023textually}. These representations are primarily categorized into two types: tokens based on speech self-supervised representations \cite{lakhotia-etal-2021-generative, hassid2023textually} and neural audio codecs \cite{wang2023neural}.

A discrete token based on speech self-supervised representation \cite{lakhotia-etal-2021-generative,lee-etal-2022-textless} is obtained by $k$-means clustering of the intermediate representation from a speech self-supervised model. These tokens, often called acoustic units, are typically encoded with a frequency range of 25Hz to 50Hz. The amount of speech information within the compressed discrete speech token is determined by the number of clusters, denoted as $k$. With a relatively small value for $k$, many works have preserved the semantic information in the tokens and utilized these to construct SLMs \cite{Polyak2021SpeechRF, communication2023seamless}. 

Neural audio codecs, another type of discrete token, capture both semantic and paralinguistic information of speech \cite{10.1109/TASLP.2021.3129994, defossez2022highfi, kumar2023highfidelity, yang2023hifi, zhang2024speechtokenizer}. A speech encoder and decoder are trained using an autoencoder architecture with residual vector quantizer for the encoder output \cite{1664069}. This representation includes most of the perceptual information of audio and is widely used for audio synthesis \cite{wang2023neural, kreuk2023audiogen, copet2023simple, agostinelli2023musiclm}.

\paragraph{Spoken Language and Dialog Models.}
Many studies have recently explored spoken language modeling to address a variety of tasks involving speech and text \cite{chu2023qwen, tang2023salmonn, wang2023blsp, team2023gemini}. Various works tackle tasks such as automatic speech recognition \cite{chen2023salm, hono2023integration, fathullah2023prompting, wang2023slm, hu2024large}, spoken question answering \cite{Pan2023COSMICDE, zhao2023librisqa, gong_ltuas}, and speech-to-text translation \cite{chen2023salm, wang2023slm, fathullah2023generalpurpose}, which process speech as input and output text. Conversely, there are also emerging works focused on tasks like speech synthesis \cite{wang2023neural, yang2023uniaudio, 10.1162/tacl_a_00618, zhang2023speak, kim2024clamtts}, where text is used as input to generate speech output.
Early SLMs that process speech as input and output are trained solely based on speech data without language models \cite{lakhotia-etal-2021-generative, nguyen-etal-2023-generative}. 
With the advancement of LLMs, several studies aim to construct SLMs that extend language models to handle both speech input and output. These studies are primarily proposed for speech modality pretraining \cite{hassid2023textually, nguyen2024spiritlm, nachmani2023spoken}, or introduced in specific tasks such as speech-to-speech translation \cite{rubenstein2023audiopalm, wang2023viola, communication2023seamlessm4t, dong2024polyvoice} and spoken conversation modeling \cite{zhang2023speechgpt, zhan2024anygpt}.

Recently, several works have been proposed for spoken dialog modeling with speech input and output \cite{nguyen-etal-2023-generative, zhang2023speechgpt, lin2024advancing}. \citet{nguyen-etal-2023-generative} develop a decoder-only transformer model trained from scratch, designed for modeling conversations between two speakers. In contrast, \citet{lin2024advancing} adopt a cascaded approach for spoken dialog modeling that consists of separate ASR, an LLM-based emotion-aware text dialog model, and emotional TTS components. \citet{zhang2023speechgpt} build SLMs on top of a pretrained LLM with objective functions designed for ASR and TTS tasks.

Among the previous works, end-to-end pipelines \cite{nguyen-etal-2023-generative, zhang2023speechgpt} that focus solely on speech-only training or leverage simple cross-modal objectives for speech-text pretraining fail to fully utilize the capabilities of pretrained language models. Additionally, cascaded models \cite{lin2024advancing}, which use separate ASR and TTS for spoken dialog, need explicit labels to incorporate paralinguistic features. This label dependency makes data collection challenging and limits the models to representing label-definable non-verbal cues. Furthermore, the error propagation inherent in the cascaded pipeline \cite{bastianelli-etal-2020-slurp} increases their susceptibility to compounded errors.

\section{Our Approach}

In this section, we describe the components that enable coherent and prosodic spoken dialog modeling, distinguishing our research from previous works. We first explain the discrete speech representation used for spoken dialog modeling in Section \ref{approach:prosody_unit}, demonstrating its suitability for prosody modeling. We then propose a unified speech-text pretraining scheme that extends the capabilities of the pretrained LLM into the domain of spoken language modeling in Section \ref{approach:unified-speech-text-pretraining}. Finally, in Section \ref{approach:unified-spoken-dialog-model} and \ref{approach:decoder}, we introduce USDM, a spoken dialog model fine-tuned with a multi-step spoken dialog template, and the speech decoder that restores the output speech token to a raw waveform.

\subsection{Speech-to-Unit Encoder}
\label{approach:prosody_unit}

To model natural speech conversations, the speech representation must contain not only the content of the speech but also paralinguistic features such as emotions, which are crucial for conversation. We adopt acoustic units as speech tokens that are derived from $k$-means clustering of a self-supervised model's intermediate speech representation, which is known to predominantly capture content and pronunciation \cite{lee-etal-2022-textless, communication2023seamlessm4t}. The information captured by an acoustic unit token varies depending on the number of clusters; the greater the number of clusters, the more encoded information. Hence, among publicly available schemes, we consider the acoustic unit tokenization scheme with the largest vocabulary size, $k=10,000$ \cite{communication2023seamlessm4t}. We then analyze whether the tokens derived from this scheme contain non-verbal information.

The acoustic unit extractor used in SeamlessM4T \cite{communication2023seamlessm4t}, first resamples the speech to a sampling rate of 16kHz and then feeds it to the XLS-R \cite{babu22_interspeech}, obtaining 50Hz intermediate continuous representations. Unit sequences are extracted by clustering these representations into 10,000 clusters, determining the vocabulary size of the speech tokens. Using this unit extractor,  we conduct two experiments to investigate features captured in the unit sequence besides the semantic content. First, we perform unit emotion recognition tasks with speech emotion recognition data to ascertain whether the unit sequences contain paralinguistic information. Next, we train a separate unit-to-speech reconstruction model and use this model to compare the original and reconstructed speech, investigating the information encoded in the unit sequences.

For the unit emotion recognition task, we train 3-layer transformer-based emotion classifiers using acoustic units on CREMA-D \cite{6849440}, which is a speech emotion recognition dataset with six emotion categories. If the units lack paralinguistic information, the classification accuracy would approximate the probability of random guessing, which is $16.6\%$. However, we observe that the classification accuracy is $60.8\%$, indicating that the acoustic units contain emotional cues.

\begin{figure}
    \centering
    \includegraphics[width=0.41\linewidth]{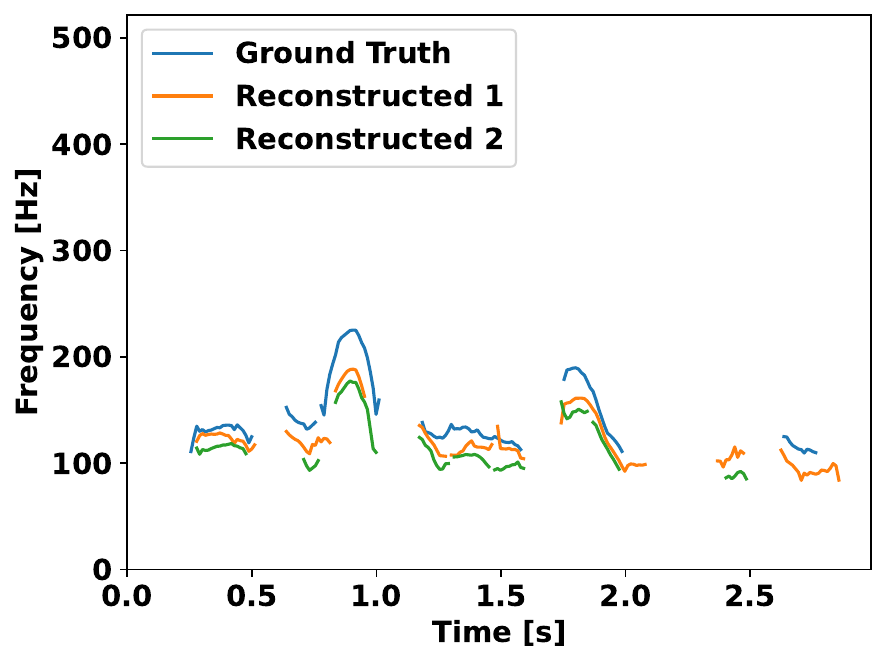} 
    \hspace{0.2cm}
    \includegraphics[width=0.41\linewidth]{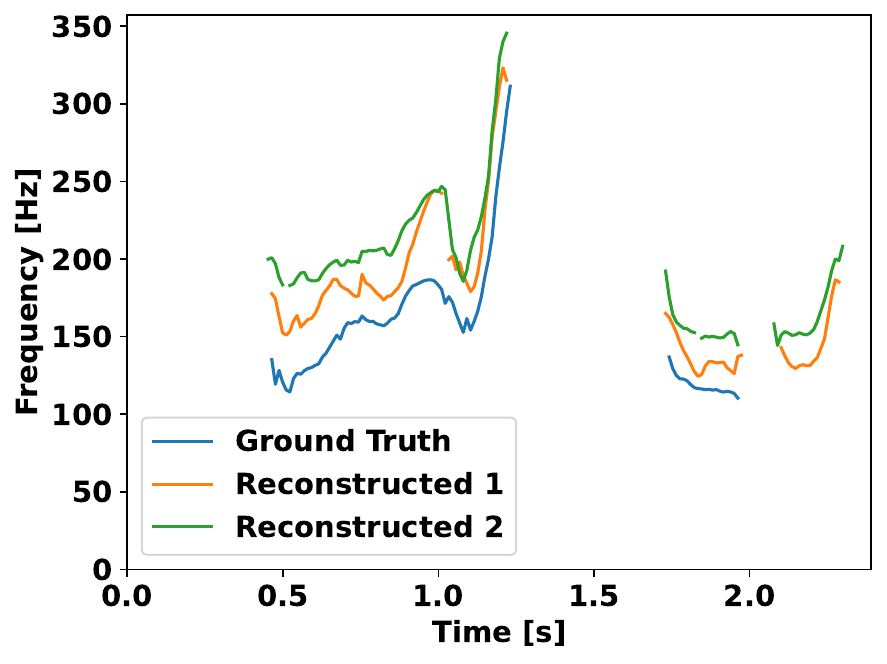} 
    \caption{Pitch contour of the original audio and the audio reconstructed from extracted acoustic units. Due to the stochastic nature of the reconstruction model, we attempt reconstruction twice, demonstrating that the pitch variation closely mirrors the ground truth.}
    \label{fig2:f0_contour}
    \vskip -0.2in
\end{figure}

To further investigate the paralinguistic information contained in the units, we train a separate unit-to-speech reconstruction model using 54,000 hours of speech data. The reconstruction model is trained using the architecture of Voicebox, one of the zero-shot stochastic TTS models \cite{le2023voicebox}. Unlike Voicebox, which takes text and reference speech as inputs for adaptation, our model is trained to generate speech solely from a unit sequence without any reference speech. The comparison between the original audio and the speech reconstructed from the extracted units shows that while the timbre and absolute pitch of the reconstructed speech differ, the pitch variation has a similar trend, as shown in Figure \ref{fig2:f0_contour}. This implies a crucial role in conveying non-verbal characteristics such as emotions, which closely match the original audio. Additionally, we have uploaded samples of several ground truth audios and corresponding reconstructed audios on our demo page.

Through these two experiments, we confirm that the acoustic units, commonly known to primarily encode semantic information, also contain a significant amount of paralinguistic information, such as emotions and pitch variations. We adopt this speech tokenization scheme for our speech-text pretraining and spoken dialog fine-tuning to help capture non-verbal cues in spoken conversations. More detailed descriptions of these experiments are provided in Appendix \ref{appendix:emotional_cue} and \ref{appendix:voicebox}.

\subsection{Unified Speech-Text Pretraining}
\label{approach:unified-speech-text-pretraining}

\begin{figure}
    \centering
    \includegraphics[width=0.9\linewidth]{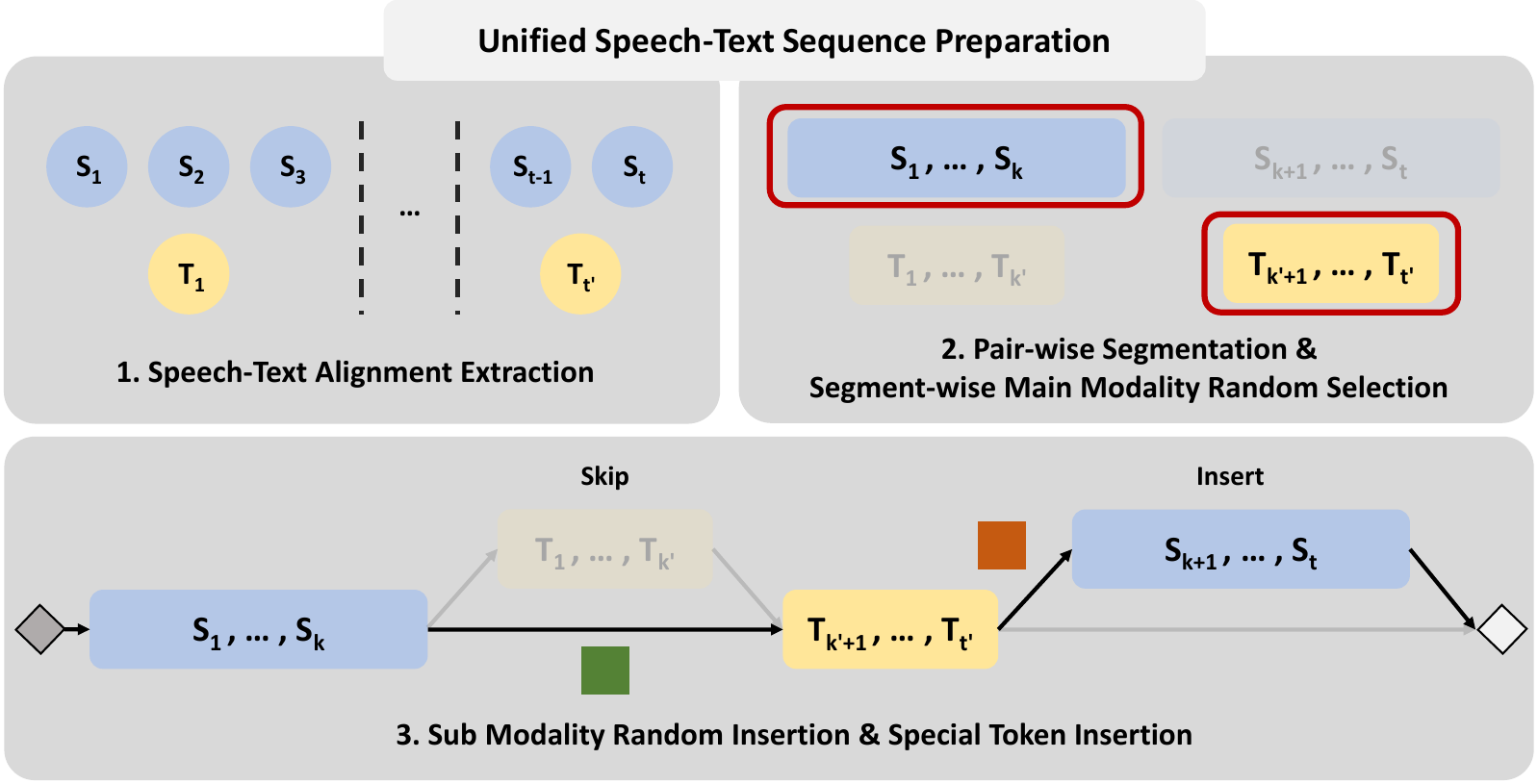}
    \includegraphics[width=0.9\linewidth]{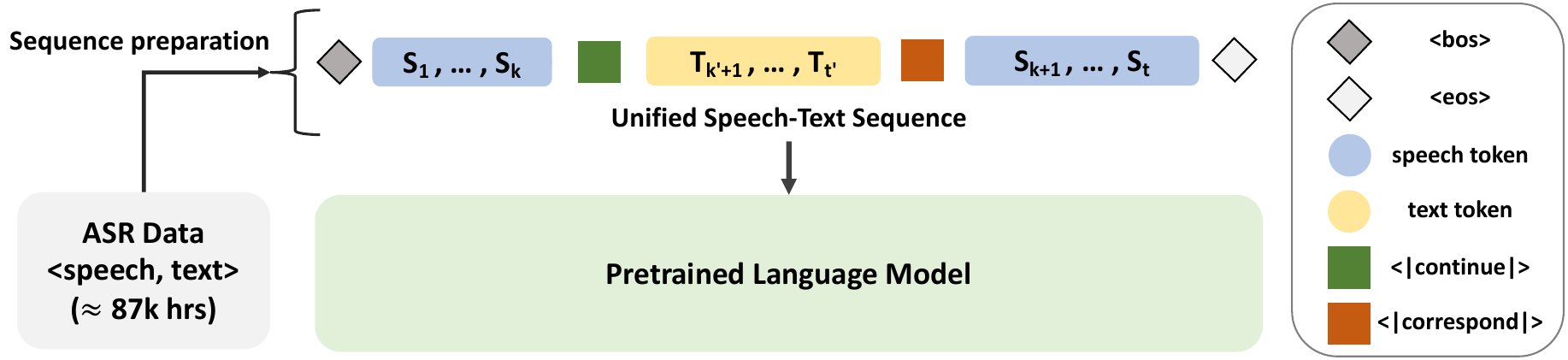}
    \caption{The overall speech-text pretraining scheme.}
    \label{fig1}
    \vskip -0.1in
\end{figure}

In this section, we introduce a unified speech-text pretraining scheme that extends pretrained LLM to speech-text cross-modality. Our overall speech-text pretraining scheme is in Figure \ref{fig1}.

For pretraining the speech-text model, we utilize Mistral-7B \cite{jiang2023mistral} as a pretrained LLM. To its existing vocabulary, we add 10,000 unit tokens and 2 special tokens, which will be described later, reinitializing only the embedding weights of these new tokens. We pretrain the speech-text model with approximately 87,000 hours of English ASR data. Each $<$speech, text$>$ pair is used to create an interleaved speech-text sample $\mathcal{I}_j=i_{1,j}, i_{2,j}, ..., i_{||\mathcal{I}_j||,j}$ to model various cross-modal relationships. Here, $i_{k,j}$ can be either an acoustic unit token, a text token, or a special token, and we construct the sequence $\mathcal{I}_j$ with a proposed per-sample speech-text interleaving method in the following paragraph. Given the dataset $\mathcal{D}=\{\mathcal{I}_1, \mathcal{I}_2, ..., \mathcal{I}_{||\mathcal{D}||}\}$, the objective of our pretraining scheme is as follows: 

\begin{equation}
\label{eq:pretraining}
    \mathcal{L}(\theta) = -\Sigma^\mathcal{||\mathcal{D}||}_{j=1}\Sigma^{||\mathcal{I}_j||}_{k=1}{\log p(i_{k,j}|i_{<k,j};\theta)},
\end{equation}

where $\theta$ refers to the parameters of LLM and the embedding weights of the newly added tokens.

When constructing a spoken language model by extending pretrained LLM, relying on a specific task, such as ASR \cite{zhang2023speechgpt}, TTS \cite{zhang2023speechgpt}, uni-modal \cite{hassid2023textually} and cross-modal continuation task \cite{wang2023blsp, nguyen2024spiritlm}, despite its large amount of dataset, may limit the model's capabilities to only those predefined relationships. To build a comprehensive speech-text model capable of both receiving and generating speech, we reinterpret the cross-modal relationship in terms of continuation or correspondence as shown in Figure \ref{fig:usdm-overview-intro}. Our proposed method, which focuses on this redefined relationship, is capable of generating diverse speech-text interleaved sequences, ensuring the model can handle complex speech-text interactions.

\textbf{Speech-Text Alignment Extraction} 
We first extract word-level alignments of the speech and its transcript using the Montreal Forced Aligner \cite{mcauliffe17_interspeech}. These alignments yield speech time intervals for each word, which we then convert into index intervals of unit sequences at a resolution of 50Hz.

\textbf{Pair-wise Segmentation and Segment-wise Main Modality Random Selection} 
Using the intervals, we divide each unit and text pair into $N$ segments. Subsequently, from each of these segments, we randomly sample data of only one modality, either unit or text. A large value of $N$ may lead to each segment containing short acoustic units and text sequences, which poses challenges in modeling unimodal text and unimodal unit sequences. To mitigate this issue, we dynamically set the value of $N$ based on the speech duration, $N = \lfloor S/10 \rfloor + 1$, where $S$ is the speech length measured in seconds.

\textbf{Sub Modality Random Insertion and Special Token Insertion} 
This segmentation and selection process allows us to generate a unified cross-modal interleaved sequence with continuation relationships. For correspondence relationship modeling, data from the non-selected modality in each segment is inserted with a $50\%$ probability after the pre-selected modality data. Additionally, to indicate the relationship between speech and text tokens, we introduce two special tokens: \texttt{<|correspond|>} and \texttt{<|continue|>}. The former indicates that a token of the corresponding remaining modality will follow, while the latter indicates that a token of the subsequent position will follow. These tokens are added to the sequence only where the modality of the data changes.

Through this procedure, we can obtain interleaved speech-text sequences $\{\mathcal{I}_j\}_{j=1,...,||\mathcal{D}||}$. These sequences enable our speech-text model to perform not only unimodal modeling but also comprehensive cross-modal modeling. These interleaved sequences are utilized in Eq. \ref{eq:pretraining} for the pretraining.

\subsection{Unified Spoken Dialog Model}
\label{approach:unified-spoken-dialog-model}

We construct the Unified Spoken Dialog Model (USDM) by fine-tuning our speech-text model with spoken dialog data, with an overview presented as shown in Figure \ref{fig:usdm-overview-intro}. The basic template for spoken dialog fine-tuning involves directly modeling the response speech tokens from the input speech tokens. However, we adopt a template designed to fully exploit the capabilities of the pretrained LLM.

Inspired by the step-by-step reasoning mechanism employed by LLMs \cite{nye2022show}, existing works \cite{rubenstein2023audiopalm, zhang2023speechgpt, nachmani2023spoken} use text to bridge the speech. Similarly, instead of directly modeling output speech from input speech, our model transcribes the speech, generates the response text, and produces the corresponding speech in an end-to-end pipeline. The insertion of the text-related tasks between speech inputs and outputs allows the model to benefit from the pretraining LLM and chained reasoning over the intermediary modality \cite{wei2022chain}. Since each stage in the pipeline attends to all input and output tokens generated in prior stages, our approach is more robust against transcription errors and better at generating contextually relevant spoken responses than if it were carried out in independent modules (i.e., the cascaded approach), which we will discuss further in Section \ref{experiments:analysis}.

The supervised fine-tuning template we use is shown in Figure \ref{fig:finetuning-template} in the Appendix. We calculate the loss using Eq. \ref{eq:pretraining} only for the input transcript, answer text, and answer unit part, as highlighted in Figure \ref{fig:finetuning-template}.

\subsection{Unit-to-Speech Decoder}
\label{approach:decoder}
We train the unit-to-speech model using the Voicebox architecture \cite{le2023voicebox} to reconstruct speech from units. Voicebox is a zero-shot TTS model that takes text and reference speech as inputs to perform personalized TTS. Unlike the reconstruction model used in Section \ref{approach:prosody_unit}, we leverage not only unit sequences but also reference speech to adapt and perform zero-shot unit-to-speech reconstruction. Our model utilizes the reference speech and the paralinguistic features contained in the units to generate prosodic spoken responses of the target speaker. For clarity, we refer to this unit-to-speech model as unit-Voicebox. More details of our decoder are in Appendix \ref{appendix:voicebox}.
\section{Experiments and Results}

\subsection{Model Comparisons}
\subsubsection{Training Details and Baselines}
\label{experiments:details}
We compare USDM to 3 baselines, From Scratch, Cascaded, and SpeechGPT \cite{zhang2023speechgpt}, on DailyTalk \cite{10095751}. DailyTalk comprises 20 hours of spoken dialog data with a sampling rate of 22,050 Hz, involving one male and one female, and we describe further details in Appendix \ref{appendix:models_datasets}.
We also present the models used for each component of USDM and the baselines in Table \ref{table:component} in the Appendix.

\textbf{USDM.} For speech-to-unit module, we adopt the official checkpoint of XLS-R \cite{babu22_interspeech} and a quantizer with $k=10,000$ \cite{communication2023seamless}, trained on 436K hours of multilingual speech data. As a speech decoder, we follow the architecture and hyperparameters of \citet{le2023voicebox} and train the unit-Voicebox on the English subset of Multilingual LibriSpeech \cite{pratap20_interspeech} and GigaSpeech \cite{GigaSpeech2021} for 10 epochs using 64 NVIDIA A100-40GB GPUs, with a batch size of 256. We use the Adam optimizer \cite{KingBa15} with a learning rate of $10^{-4}$. We utilize the official checkpoint of BigVGAN \cite{lee2023bigvgan} as our vocoder.

Our proposed unified speech-text pretraining is conducted using 512 NVIDIA A100-40GB GPUs, with a global batch size of 1,024 for 8,000 iterations. 
For pretraining, we utilize approximately 87,000 hours of English ASR data; the English subset of Multilingual LibriSpeech \cite{pratap20_interspeech}, People's Speech \cite{galvez2021the}, GigaSpeech \cite{GigaSpeech2021}, Common Voice 15.0 \cite{ardila-etal-2020-common}, and the English subset of Voxpopuli \cite{wang-etal-2021-voxpopuli}. 
The data used for pretraining is packed to a maximum sequence length of 8,192. 
For spoken dialog modeling, we fine-tune a speech-text model with a global batch size of 64 for 5 epochs. We use linear learning rate scheduling with a peak learning rate of $2\cdot10^{-5}$ for both pretraining and fine-tuning.

\textbf{From Scratch.} The From Scratch model is nearly identical to the USDM but excludes speech-text pretraining. Specifically, we fine-tune the pretrained Mistral-7B model directly on spoken dialog data, with the hyperparameters identical to those of the USDM.

\textbf{Cascaded.} We include a Cascaded model, which employs separate ASR and TTS models, as a baseline for comparison. For the ASR model, we use the official checkpoint of \textit{whisper-large-v3} \cite{pmlr-v202-radford23a}, which is trained on 5M hours of speech data. For the speech synthesis model, we train Voicebox with text input using the same hyperparameters and datasets as unit-Voicebox. As the LLM, we utilize the transcript of the spoken dialog dataset to create text dialog data and fine-tune the Mistral-7B on this data using the same hyperparameters as the USDM.

\textbf{SpeechGPT.} For SpeechGPT \cite{zhang2023speechgpt}, we use the official implementations and checkpoints for the speech-to-unit module, spoken language model, and speech decoder module.
Specifically, we fine-tune \textit{SpeechGPT-7B-cm}, a pretrained speech-text model, with DailyTalk for a fair comparison. 

\subsubsection{Evaluation and Comparison Results}
\label{experiments:evaluation}
We conduct various evaluations on the spoken responses generated for the given spoken dialogs. When generating samples for evaluation, we adopt a sampling scheme with top\_k $=40$, top\_p $=0.7$, and temperature $=0.3$, except for SpeechGPT, where we use their own strategy. For Voicebox and unit-Voicebox, we utilize the speech from the previous turn as the reference speech. While SpeechGPT generates audio at 16kHz, other models synthesize speech at 22,050Hz. For a fair comparison, all audio samples are resampled to 16kHz and volume normalized to -27dB for evaluation.

To compare the overall preference of our model and the baselines, we conduct a human preference test via Amazon Mechanical Turk. Given a randomly selected 50 spoken dialogs from the test split of a dataset, we instruct the evaluators to compare the spoken response of our model and the baseline, considering the comprehensive aspects such as naturalness, prosody, and semantic coherence. To evaluate the prosody and the naturalness, we additionally measure the 5-scale prosody mean opinion score (P-MOS) and the 5-scale mean opinion score (MOS) through Amazon Mechanical Turk. As explained in Section \ref{approach:unified-spoken-dialog-model}, our model first generates the text to be spoken before generating a spoken response, which allows us to fix the content of the generated speech by predetermining the text. Unlike the aforementioned human preference test, to focus solely on the prosody and naturalness, respectively, we provide the ground truth response text to the model to ensure consistency in the content of the output speech, thereby preventing difficulties in evaluations that may arise from variations in content. Instructions and detailed descriptions of our evaluations are in Appendix \ref{appendix:mos}.

Furthermore, to evaluate the content appropriateness of spoken responses, we first generate the spoken responses of all models given the spoken dialogs for the entire test set. We then pass these samples through the ASR model, \textit{whisper-large-v3}, to calculate METEOR and ROUGE-L scores, which are widely used in various NLP tasks such as text summarization and are commonly employed to measure performance in dialog modeling \cite{10.1162/tacl_a_00347, zhang2019dialogpt}. We also conduct a GPT-4-based \cite{openai2023gpt4} preference test \cite{zheng2023judging} between the transcribed texts of our model and all baselines.

\begin{table}
\caption{Human evaluation results of our model and the baselines. We report the MOS and P-MOS scores with a 95\% confidence interval.}
\label{table:model_comparion}
\small
\begin{center}
\begin{tabular}{l|ccc|cc}
\toprule
\multirow{2}{*}{\textbf{Method}} & \multicolumn{3}{c|}{\textbf{Overall}}                             & \multicolumn{2}{c}{\textbf{Acoustic}} \\ \cmidrule{2-6} 
                                 & \textbf{\textit{win}} & \textbf{\textit{tie}} & \textbf{\textit{lose}} & \textbf{MOS}      & \textbf{P-MOS}    \\ \midrule
Ground Truth                     & $45.9\%$             & $8.0\%$               & $46.1\%$               & $4.51\pm0.05$     & $4.35\pm0.05$     \\
USDM                             & $-$                  & $-$                   & $-$                    & $4.31\pm0.07$     & $4.31\pm0.06$     \\
Cascaded                         & $55.3\%$             & $4.9\%$               & $39.8\%$               & $4.26\pm0.07$     & $4.22\pm0.07$     \\
From Scratch                     & $53.3\%$             & $7.6\%$               & $39.1\%$               & $3.71\pm0.11$     & $3.65\pm0.10$     \\
SpeechGPT \cite{zhang2023speechgpt}                        & $53.8\%$             & $6.9\%$               & $39.3\%$               & $4.08\pm0.09$     & $4.04\pm0.08$     \\
\bottomrule
\end{tabular}
\vskip -0.5in
\end{center}
\vskip -0.2in
\end{table}

\begin{table}[]
\caption{GPT-4 evaluation and quantitative results of our model and the baselines.}
\label{table:model_comparion_quantitative}
\small
\begin{center}
\begin{tabular}{l|ccccc|cc}
\toprule
\multirow{2}{*}{\textbf{Method}} & \multicolumn{5}{c|}{\textbf{Semantic}}                             & \multicolumn{2}{c}{\textbf{WER}}            \\ \cmidrule{2-8} 
                                 & \textbf{\textit{win}} & \textbf{\textit{tie}} & \textbf{\textit{lose}} & \multicolumn{1}{|c}{\textbf{METEOR}} & \textbf{ROUGE-L} & \textbf{STT}   & \textbf{TTS}   \\ \midrule
Ground Truth                     & $32.7\%$             & $19.6\%$              & $47.7\%$               & \multicolumn{1}{|c}{$-$}             & $-$              & $-$              & $2.2\%$              \\
USDM                             & $-$                  & $-$                   & $-$                    & \multicolumn{1}{|c}{$13.1$}          & $15.7$           & $7.4\%$              & $2.0\%$              \\
Cascaded                         & $42.7\%$             & $24.6\%$              & $32.7\%$               & \multicolumn{1}{|c}{$12.5$}          & $15.0$           & $3.8\%$              & $1.3\%$              \\
From Scratch                     & $79.7\%$             & $10.1\%$              & $10.2\%$               & \multicolumn{1}{|c}{$8.6$}           & $10.6$           & $58.1\%$             & $64.0\%$                \\
SpeechGPT \cite{zhang2023speechgpt}                        & $61.0\%$             & $13.1\%$              & $25.9\%$               & \multicolumn{1}{|c}{$9.9$}           & $11.8$           & $12.4\%$             & $23.2\%$             \\
\bottomrule
\end{tabular}
\vskip -0.5in
\end{center}
\vskip -0.2in
\end{table}

We also measure the Word Error Rate (WER) for the speech-to-text part and text-to-speech part of each model. For the speech-to-text part (STT WER), we calculate the WER across the entire test set. We use the outputs from the \textit{whisper-large-v3} model for the Cascaded baseline, while the remaining end-to-end pipelines are assessed using the intermediate transcribed text of each model. For the text-to-speech part (TTS WER), similar to our MOS and P-MOS evaluations, we calculate the WER using generated samples given the randomly selected 50 spoken dialogs and the corresponding ground truth written-form response. We generate each spoken response 5 times and report the average WER. For measuring TTS WER, we utilize the \textit{whisper-large-v3} model as the ASR model.

The results are presented in Table \ref{table:model_comparion} and \ref{table:model_comparion_quantitative}. In human preference tests that consider comprehensive factors, our model is preferred similarly to the Ground Truth and demonstrates superior preferences compared to the baselines ($p$-value $< 0.05$ from the Wilcoxon signed-rank test). For the semantic aspect, our USDM outperforms the baselines in both quantitative evaluations and the GPT-4-based preference test ($p$-value $< 0.05$). We also observe that our model surpasses the baselines in the P-MOS evaluations ($p$-value $< 0.05$), which measure the prosody naturalness of the speech given spoken dialog. Notably, the USDM shows superior prosody compared to the Cascaded model.
These results demonstrate that our model effectively incorporates prosody information in the spoken language model and is capable of generating spoken responses with content well-aligned to input speech. 

We also confirm that cross-modal pretraining is essential to leverage the capabilities of LLM. We observe that the From Scratch model, which directly models spoken dialog without pretraining, tends to overlook the bridging text and generates a spoken response that does not correspond to the pre-generated written response, thus negatively impacting its performance. This results in worse TTS and STT WERs and adversely affects the P-MOS and MOS, which are based on the prosody and naturalness of the spoken response given the transcript. This result indicates the difficulty of transferring the capabilities of text models to spoken dialog modeling without cross-modal pretraining. 

\subsection{Ablation Studies}
\subsubsection{Pretraining Schemes}
\label{experiments:pretraining}
In this section, we compare the effects of correspondence and continuation modeling, which are crucial to our pretraining method. We consider three additional pretraining schemes. \textbf{Setup 1} uses an interleaved unit-text sequence without a correspondence relationship, relying solely on continuation, similar to previous works \cite{wang2023blsp, nguyen2024spiritlm}. \textbf{Setup 2} utilizes data that maintains only a correspondence relationship, as seen in \citet{zhang2023speechgpt}. \textbf{Setup 3} is similar to our fine-tuning approach, interleaving speech with its transcript and subsequent text and speech, as proposed in \citet{nachmani2023spoken}. All setups are trained in the same way as our speech-text pretraining, with details in Section \ref{appendix:ablations}.

We evaluate these pretrained models on sequence modeling and spoken dialog modeling tasks. Performance is first assessed by measuring perplexity (PPL) of various speech-text sequences from the \texttt{test-clean} subset of the LibriSpeech dataset \cite{7178964}.  We construct six types of interleaved sequences: unimodal sequences for both unit (1) and text (2), sequences with unit followed by their corresponding text (3) and vice versa (4) to evaluate correspondence relationships, and sequences generated by dividing the unit and text in half, combining the first half's unit with the remaining half's text (5), and the first half's text with the remaining unit (6) for assessing continuation. We then calculate the average PPL for all combinations by taking the logarithm of each subsequent modality's PPL within each sequence type, averaging these logarithmic values, and then applying the exponential function. For spoken dialog modeling, we fine-tune each model with DailyTalk and measure STT WER, TTS WER, METEOR, and ROUGE-L, as described in Section \ref{experiments:evaluation}.

\begin{table}[]
\caption{Results of the ablation studies on the pretraining and fine-tuning schemes. For PPL, we report the average PPL for each modality across the six combinations described in the text.}
\label{table:ablation}
\small
\begin{center}
\begin{tabular}{l|cc|cccc}
\toprule
\multicolumn{1}{c|}{\multirow{2}{*}{\textbf{Method}}} & \multicolumn{2}{c|}{\textbf{Pretraining}}             & \multicolumn{4}{c}{\textbf{Spoken Dialog Modeling}}                                           \\ \cmidrule{2-7} 
\multicolumn{1}{c|}{}                                 & \textbf{Text PPL} & \textbf{Unit PPL} & \textbf{STT WER} & \multicolumn{1}{c|}{\textbf{TTS WER}} & \textbf{METEOR} & \textbf{ROUGE-L} \\ \midrule
Ours                                                  & $6.886$                   & $4.813$                   & $7.4\%$          & \multicolumn{1}{c|}{$2.0\%$}                 & $13.1$          & $15.7$           \\
Setup 1                                               & $14.485$                  & $5.261$                   & $57.8\%$         & \multicolumn{1}{c|}{$82.1\%$}                 & $8.9$           & $10.6$           \\
Setup 2                                               & $31.679$                  & $5.619$                   & $11.2\%$         & \multicolumn{1}{c|}{$2.5\%$}                 & $12.5$          & $15.1$           \\
Setup 3                                               & $21.392$                  & $5.146$                   & $7.3\%$          & \multicolumn{1}{c|}{$2.0\%$}                 & $12.7$          & $15.4$           \\ \midrule
S1 $\rightarrow$ S2                                              & $-$                       & $-$                       & $-$              & \multicolumn{1}{c|}{$-$}                 & $6.5$           & $7.7$      \\ 
\bottomrule
\end{tabular}
\vskip -0.5in
\end{center}
\vskip -0.2in
\end{table}

We present the average PPL of each modality in Table \ref{table:ablation} and the PPL for each combination in Table \ref{table:pretrained_ppl} in the Appendix. Our model demonstrates superior average PPL across both modalities. Focusing solely on either correspondence or continuation relationships, or on specific templates tends to make the model specialize in certain objectives but hinders its ability to model diverse relationships effectively. Our proposed unified speech-text pretraining scheme performs uniformly well without being overly focused on specific relationships. We also show in Table \ref{table:ablation} that our speech-text model is beneficial to spoken dialog modeling, as evidenced by the WER, METEOR, and ROUGE-L scores. Particularly, Setup 1, which lacks correspondence relationship pretraining, exhibits significantly higher STT and TTS WERs, resulting in deteriorated semantic performance in spoken responses. 

\subsubsection{Fine-tuning Schemes}
As explained in Section \ref{approach:unified-spoken-dialog-model}, USDM first models the input and output text as a bridge when given a speech input before generating the spoken response. To demonstrate this approach, we train a spoken dialog model that models the speech output directly from the speech input (S1 $\rightarrow$ S2). We evaluate the generated response speech through METEOR and ROUGE-L scores with the same samples described in Section \ref{experiments:evaluation}, and the results are shown in Table \ref{table:ablation}. We find that intermediate text modeling in spoken dialog generation helps generate appropriate spoken responses. 
This confirms that the process of generating text before speech leverages the capabilities of the pretrained model effectively.

\subsection{Analysis on Input Modality}
\label{experiments:analysis}

As seen in Table \ref{table:model_comparion_quantitative}, the Cascaded model exhibits a lower ASR WER compared to USDM. This is due to the separate ASR model used in the Cascaded model, \textit{whisper-large-v3}, which has been trained on approximately 5 million hours of speech data. However, in terms of the semantics of the spoken response, our model outperforms the Cascaded model. 

\begin{figure}
    \centering
    \includegraphics[width=0.8\linewidth]{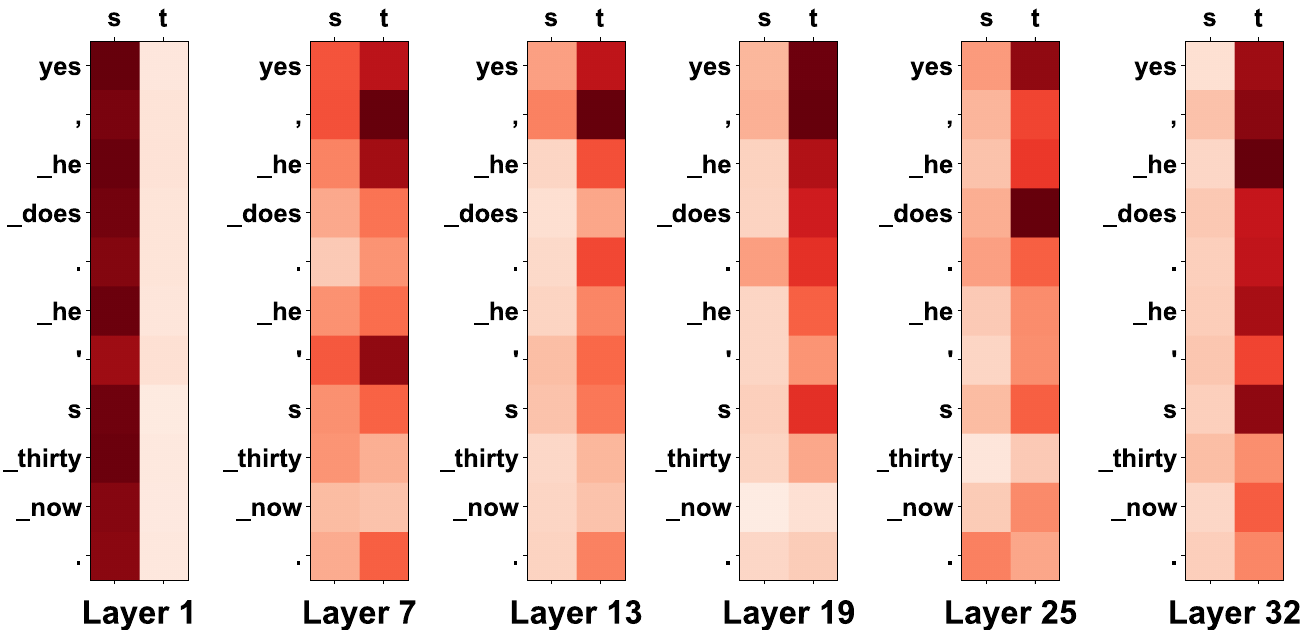}
    \caption{Attention maps between the generated responses of the USDM and the input speech (s) and its transcribed text (t). Input speech: ``Oh, I can't believe it. He looks very young.''}
    \label{fig3}
    \vskip -0.1in
\end{figure}

Similar to previous works that show the advantages of end-to-end pipelines over cascaded approaches in several tasks \cite{10446782, 10446635, wang23ga_interspeech, Xue2022LargeScaleSE}, USDM leverages input speech to generate more semantically coherent answers. To empirically verify that our generated text responses utilize information from both the preceding transcript and the input speech, we plot the attention maps for each layer, as illustrated in Figure \ref{fig3}. We calculate the probability that each token in the generated response attends to each token in the input unit sequence and corresponding transcript by averaging the probabilities across all heads of the attention modules for each layer. Subsequently, we aggregate these probabilities for input unit tokens to compute a cumulative probability for the speech input, and similarly for text tokens relative to the transcript. As shown in Figure \ref{fig3}, our generated tokens attend not only to the transcribed text but also to the speech input, indicating that our model benefits from the speech input.

In Figure \ref{fig3}, we also observe that the generated responses notably attend to the transcribed text. To analyze the impact of more accurate transcription on model performance, we substitute the model-generated input transcript with the ground truth transcript in the middle of the inference of USDM and measure the METEOR and ROUGE-L scores for the generated spoken responses. The measured scores are 13.6 and 16.2, respectively, surpassing our model's previous results of 13.1 and 15.7 in Table \ref{table:model_comparion_quantitative}. This confirms that enhancing the accuracy of unit-to-text conversion in USDM also improves the semantic coherence of the spoken responses.

\section{Conclusion}

In this work, we presented USDM, a model synthesizing spoken dialog responses enriched with natural prosody. 
We proposed a novel speech-text pretraining scheme that models the comprehensive relationship between speech and text, which proves beneficial for spoken dialog modeling.
Our approach is complemented by leveraging an acoustic unit tokenization scheme that preserves prosodic information, coupled with a supporting pair of an encoder and a decoder. 
We showed that USDM outperforms the baselines regarding content, prosody, and naturalness as a spoken response for the DailyTalk dataset.
Additionally, we demonstrated that our pretraining and fine-tuning scheme benefits the USDM in modeling spoken dialog through ablation studies.
Various samples for diverse scenarios in our demo page also showcased the capabilities of USDM.
We believe that USDM has laid the groundwork for extending the conversational capabilities of LLMs to the voice domains.

Despite these advantages, our model has several limitations and areas for improvement. Firstly, the exploration of datasets and models used for pretraining is limited. Further investigation is necessary to determine which data are crucial for our pretraining scheme and to explore whether our pretraining scheme could be effective with other LLMs beyond Mistral-7B. Secondly, building a spoken dialog model capable of directly generating spoken responses from input spoken dialog without the need for cross-modal chaining can be a promising direction. Next, the current pretraining scheme is based on tens of thousands of hours of English data, and it has limitations when applied to other languages with relatively smaller amounts of speech data compared to English. We plan to expand our model to a variety of languages in addition to English. Lastly, we also plan to investigate whether our pretraining approach is beneficial for other speech-text tasks beyond spoken dialog modeling. 

\begin{ack}
This work was supported by SNU-Naver Hyperscale AI Center, Institute of Information \& communications Technology Planning \& Evaluation (IITP) grant funded by the Korea government(MSIT) [NO.RS-2021-II211343, Artificial Intelligence Graduate School Program (Seoul National University)], National Research Foundation of Korea (NRF) grant funded by the Korea government (MSIT) (No. 2022R1A3B1077720, No.2022R1A5A7083908), and the BK21 FOUR program of the Education and Research Program for Future ICT Pioneers, SNU in 2024.
\end{ack}

\newpage
\bibliography{main}

\begin{thebibliography}{90}
\providecommand{\natexlab}[1]{#1}
\providecommand{\url}[1]{\texttt{#1}}
\expandafter\ifx\csname urlstyle\endcsname\relax
  \providecommand{\doi}[1]{doi: #1}\else
  \providecommand{\doi}{doi: \begingroup \urlstyle{rm}\Url}\fi

\bibitem[OpenAI(2023)]{openai2023gpt4}
OpenAI.
\newblock {GPT-4} technical report.
\newblock \emph{CoRR}, abs/2303.08774, 2023.
\newblock \doi{10.48550/ARXIV.2303.08774}.
\newblock URL \url{https://doi.org/10.48550/arXiv.2303.08774}.

\bibitem[Wei et~al.(2022{\natexlab{a}})Wei, Tay, Bommasani, Raffel, Zoph, Borgeaud, Yogatama, Bosma, Zhou, Metzler, Chi, Hashimoto, Vinyals, Liang, Dean, and Fedus]{wei2022emergent}
Jason Wei, Yi~Tay, Rishi Bommasani, Colin Raffel, Barret Zoph, Sebastian Borgeaud, Dani Yogatama, Maarten Bosma, Denny Zhou, Donald Metzler, Ed~H. Chi, Tatsunori Hashimoto, Oriol Vinyals, Percy Liang, Jeff Dean, and William Fedus.
\newblock Emergent abilities of large language models.
\newblock \emph{Transactions on Machine Learning Research}, 2022{\natexlab{a}}.
\newblock ISSN 2835-8856.
\newblock URL \url{https://openreview.net/forum?id=yzkSU5zdwD}.
\newblock Survey Certification.

\bibitem[Brown et~al.(2020)Brown, Mann, Ryder, Subbiah, Kaplan, Dhariwal, Neelakantan, Shyam, Sastry, Askell, et~al.]{brown2020language}
Tom~B Brown, Benjamin Mann, Nick Ryder, Melanie Subbiah, Jared Kaplan, Prafulla Dhariwal, Arvind Neelakantan, Pranav Shyam, Girish Sastry, Amanda Askell, et~al.
\newblock Language models are few-shot learners.
\newblock In \emph{Proceedings of the 34th International Conference on Neural Information Processing Systems}, pages 1877--1901, 2020.

\bibitem[Kaplan et~al.(2020)Kaplan, McCandlish, Henighan, Brown, Chess, Child, Gray, Radford, Wu, and Amodei]{kaplan2020scaling}
Jared Kaplan, Sam McCandlish, Tom Henighan, Tom~B Brown, Benjamin Chess, Rewon Child, Scott Gray, Alec Radford, Jeffrey Wu, and Dario Amodei.
\newblock Scaling laws for neural language models.
\newblock \emph{arXiv preprint arXiv:2001.08361}, 2020.

\bibitem[Team(2024)]{yoo2024hyperclovaxtechnicalreport}
HyperCLOVA~X Team.
\newblock Hyperclova x technical report, 2024.
\newblock URL \url{https://arxiv.org/abs/2404.01954}.

\bibitem[Wei et~al.(2022{\natexlab{b}})Wei, Wang, Schuurmans, Bosma, Xia, Chi, Le, Zhou, et~al.]{wei2022chain}
Jason Wei, Xuezhi Wang, Dale Schuurmans, Maarten Bosma, Fei Xia, Ed~Chi, Quoc~V Le, Denny Zhou, et~al.
\newblock Chain-of-thought prompting elicits reasoning in large language models.
\newblock \emph{Advances in Neural Information Processing Systems}, 35:\penalty0 24824--24837, 2022{\natexlab{b}}.

\bibitem[Yao et~al.(2022)Yao, Zhao, Yu, Du, Shafran, Narasimhan, and Cao]{yao2022react}
Shunyu Yao, Jeffrey Zhao, Dian Yu, Nan Du, Izhak Shafran, Karthik~R Narasimhan, and Yuan Cao.
\newblock React: Synergizing reasoning and acting in language models.
\newblock In \emph{The Eleventh International Conference on Learning Representations}, 2022.

\bibitem[Ouyang et~al.(2022)Ouyang, Wu, Jiang, Almeida, Wainwright, Mishkin, Zhang, Agarwal, Slama, Ray, et~al.]{ouyang2022training}
Long Ouyang, Jeffrey Wu, Xu~Jiang, Diogo Almeida, Carroll Wainwright, Pamela Mishkin, Chong Zhang, Sandhini Agarwal, Katarina Slama, Alex Ray, et~al.
\newblock Training language models to follow instructions with human feedback.
\newblock \emph{Advances in Neural Information Processing Systems}, 35:\penalty0 27730--27744, 2022.

\bibitem[Gabriel et~al.(2024)Gabriel, Manzini, Keeling, Hendricks, Rieser, Iqbal, Tomašev, Ktena, Kenton, Rodriguez, El-Sayed, Brown, Akbulut, Trask, Hughes, Bergman, Shelby, Marchal, Griffin, Mateos-Garcia, Weidinger, Street, Lange, Ingerman, Lentz, Enger, Barakat, Krakovna, Siy, Kurth-Nelson, McCroskery, Bolina, Law, Shanahan, Alberts, Balle, de~Haas, Ibitoye, Dafoe, Goldberg, Krier, Reese, Witherspoon, Hawkins, Rauh, Wallace, Franklin, Goldstein, Lehman, Klenk, Vallor, Biles, Morris, King, y~Arcas, Isaac, and Manyika]{gabriel2024ethics}
Iason Gabriel, Arianna Manzini, Geoff Keeling, Lisa~Anne Hendricks, Verena Rieser, Hasan Iqbal, Nenad Tomašev, Ira Ktena, Zachary Kenton, Mikel Rodriguez, Seliem El-Sayed, Sasha Brown, Canfer Akbulut, Andrew Trask, Edward Hughes, A.~Stevie Bergman, Renee Shelby, Nahema Marchal, Conor Griffin, Juan Mateos-Garcia, Laura Weidinger, Winnie Street, Benjamin Lange, Alex Ingerman, Alison Lentz, Reed Enger, Andrew Barakat, Victoria Krakovna, John~Oliver Siy, Zeb Kurth-Nelson, Amanda McCroskery, Vijay Bolina, Harry Law, Murray Shanahan, Lize Alberts, Borja Balle, Sarah de~Haas, Yetunde Ibitoye, Allan Dafoe, Beth Goldberg, Sébastien Krier, Alexander Reese, Sims Witherspoon, Will Hawkins, Maribeth Rauh, Don Wallace, Matija Franklin, Josh~A. Goldstein, Joel Lehman, Michael Klenk, Shannon Vallor, Courtney Biles, Meredith~Ringel Morris, Helen King, Blaise~Agüera y~Arcas, William Isaac, and James Manyika.
\newblock The ethics of advanced ai assistants, 2024.

\bibitem[Horowitz and Samuels(1987)]{horowitz1987comprehending}
Rosalind Horowitz and S~Jay Samuels.
\newblock Comprehending oral and written language: Critical contrasts for literacy and schooling.
\newblock In \emph{Comprehending oral and written language}, pages 1--52. Brill, 1987.

\bibitem[Clark et~al.(2019)Clark, Doyle, Garaialde, Gilmartin, Schl{\"o}gl, Edlund, Aylett, Cabral, Munteanu, Edwards, et~al.]{clark2019state}
Leigh Clark, Philip Doyle, Diego Garaialde, Emer Gilmartin, Stephan Schl{\"o}gl, Jens Edlund, Matthew Aylett, Jo{\~a}o Cabral, Cosmin Munteanu, Justin Edwards, et~al.
\newblock The state of speech in hci: Trends, themes and challenges.
\newblock \emph{Interacting with computers}, 31\penalty0 (4):\penalty0 349--371, 2019.

\bibitem[Bommasani et~al.(2021)Bommasani, Hudson, Adeli, Altman, Arora, von Arx, Bernstein, Bohg, Bosselut, Brunskill, et~al.]{bommasani2021opportunities}
Rishi Bommasani, Drew~A Hudson, Ehsan Adeli, Russ Altman, Simran Arora, Sydney von Arx, Michael~S Bernstein, Jeannette Bohg, Antoine Bosselut, Emma Brunskill, et~al.
\newblock On the opportunities and risks of foundation models.
\newblock \emph{arXiv preprint arXiv:2108.07258}, 2021.

\bibitem[Tsimpoukelli et~al.(2021)Tsimpoukelli, Menick, Cabi, Eslami, Vinyals, and Hill]{tsimpoukelli2021multimodal}
Maria Tsimpoukelli, Jacob~L Menick, Serkan Cabi, SM~Eslami, Oriol Vinyals, and Felix Hill.
\newblock Multimodal few-shot learning with frozen language models.
\newblock \emph{Advances in Neural Information Processing Systems}, 34:\penalty0 200--212, 2021.

\bibitem[Alayrac et~al.(2022)Alayrac, Donahue, Luc, Miech, Barr, Hasson, Lenc, Mensch, Millican, Reynolds, et~al.]{alayrac2022flamingo}
Jean-Baptiste Alayrac, Jeff Donahue, Pauline Luc, Antoine Miech, Iain Barr, Yana Hasson, Karel Lenc, Arthur Mensch, Katherine Millican, Malcolm Reynolds, et~al.
\newblock Flamingo: a visual language model for few-shot learning.
\newblock \emph{Advances in Neural Information Processing Systems}, 35:\penalty0 23716--23736, 2022.

\bibitem[Cho et~al.(2021)Cho, Lei, Tan, and Bansal]{cho2021unifying}
Jaemin Cho, Jie Lei, Hao Tan, and Mohit Bansal.
\newblock Unifying vision-and-language tasks via text generation.
\newblock In \emph{International Conference on Machine Learning}, pages 1931--1942. PMLR, 2021.

\bibitem[Yu et~al.(2023)Yu, Shi, Pasunuru, Muller, Golovneva, Wang, Babu, Tang, Karrer, Sheynin, et~al.]{yu2023scaling}
Lili Yu, Bowen Shi, Ramakanth Pasunuru, Benjamin Muller, Olga Golovneva, Tianlu Wang, Arun Babu, Binh Tang, Brian Karrer, Shelly Sheynin, et~al.
\newblock Scaling autoregressive multi-modal models: Pretraining and instruction tuning.
\newblock \emph{arXiv preprint arXiv:2309.02591}, 2023.

\bibitem[Lakhotia et~al.(2021)Lakhotia, Kharitonov, Hsu, Adi, Polyak, Bolte, Nguyen, Copet, Baevski, Mohamed, and Dupoux]{lakhotia-etal-2021-generative}
Kushal Lakhotia, Eugene Kharitonov, Wei-Ning Hsu, Yossi Adi, Adam Polyak, Benjamin Bolte, Tu-Anh Nguyen, Jade Copet, Alexei Baevski, Abdelrahman Mohamed, and Emmanuel Dupoux.
\newblock On generative spoken language modeling from raw audio.
\newblock \emph{Transactions of the Association for Computational Linguistics}, 9:\penalty0 1336--1354, 2021.
\newblock \doi{10.1162/tacl_a_00430}.
\newblock URL \url{https://aclanthology.org/2021.tacl-1.79}.

\bibitem[Chen et~al.(2022)Chen, Wang, Chen, Wu, Liu, Chen, Li, Kanda, Yoshioka, Xiao, et~al.]{chen2022wavlm}
Sanyuan Chen, Chengyi Wang, Zhengyang Chen, Yu~Wu, Shujie Liu, Zhuo Chen, Jinyu Li, Naoyuki Kanda, Takuya Yoshioka, Xiong Xiao, et~al.
\newblock Wavlm: Large-scale self-supervised pre-training for full stack speech processing.
\newblock \emph{IEEE Journal of Selected Topics in Signal Processing}, 16\penalty0 (6):\penalty0 1505--1518, 2022.

\bibitem[Nguyen et~al.(2023{\natexlab{a}})Nguyen, Kharitonov, Copet, Adi, Hsu, Elkahky, Tomasello, Algayres, Sagot, Mohamed, and Dupoux]{nguyen-etal-2023-generative}
Tu~Anh Nguyen, Eugene Kharitonov, Jade Copet, Yossi Adi, Wei-Ning Hsu, Ali Elkahky, Paden Tomasello, Robin Algayres, Beno{\^\i}t Sagot, Abdelrahman Mohamed, and Emmanuel Dupoux.
\newblock Generative spoken dialogue language modeling.
\newblock \emph{Transactions of the Association for Computational Linguistics}, 11:\penalty0 250--266, 2023{\natexlab{a}}.
\newblock \doi{10.1162/tacl_a_00545}.
\newblock URL \url{https://aclanthology.org/2023.tacl-1.15}.

\bibitem[Wang et~al.(2023{\natexlab{a}})Wang, Chen, Wu, Zhang, Zhou, Liu, Chen, Liu, Wang, Li, He, Zhao, and Wei]{wang2023neural}
Chengyi Wang, Sanyuan Chen, Yu~Wu, Ziqiang Zhang, Long Zhou, Shujie Liu, Zhuo Chen, Yanqing Liu, Huaming Wang, Jinyu Li, Lei He, Sheng Zhao, and Furu Wei.
\newblock Neural codec language models are zero-shot text to speech synthesizers, 2023{\natexlab{a}}.

\bibitem[Arjmand et~al.(2021)Arjmand, Dousti, and Moradi]{arjmand2021teasel}
Mehdi Arjmand, Mohammad~Javad Dousti, and Hadi Moradi.
\newblock Teasel: a transformer-based speech-prefixed language model.
\newblock \emph{arXiv preprint arXiv:2109.05522}, 2021.

\bibitem[Ao et~al.(2021)Ao, Wang, Zhou, Wang, Ren, Wu, Liu, Ko, Li, Zhang, et~al.]{ao2021speecht5}
Junyi Ao, Rui Wang, Long Zhou, Chengyi Wang, Shuo Ren, Yu~Wu, Shujie Liu, Tom Ko, Qing Li, Yu~Zhang, et~al.
\newblock Speecht5: Unified-modal encoder-decoder pre-training for spoken language processing.
\newblock \emph{arXiv preprint arXiv:2110.07205}, 2021.

\bibitem[Chu et~al.(2023)Chu, Xu, Zhou, Yang, Zhang, Yan, Zhou, and Zhou]{chu2023qwen}
Yunfei Chu, Jin Xu, Xiaohuan Zhou, Qian Yang, Shiliang Zhang, Zhijie Yan, Chang Zhou, and Jingren Zhou.
\newblock Qwen-audio: Advancing universal audio understanding via unified large-scale audio-language models.
\newblock \emph{arXiv preprint arXiv:2311.07919}, 2023.

\bibitem[Rubenstein et~al.(2023)Rubenstein, Asawaroengchai, Nguyen, Bapna, Borsos, Quitry, Chen, Badawy, Han, Kharitonov, et~al.]{rubenstein2023audiopalm}
Paul~K Rubenstein, Chulayuth Asawaroengchai, Duc~Dung Nguyen, Ankur Bapna, Zal{\'a}n Borsos, F{\'e}lix de~Chaumont Quitry, Peter Chen, Dalia~El Badawy, Wei Han, Eugene Kharitonov, et~al.
\newblock Audiopalm: A large language model that can speak and listen.
\newblock \emph{arXiv preprint arXiv:2306.12925}, 2023.

\bibitem[Zhang et~al.(2023{\natexlab{a}})Zhang, Li, Zhang, Zhan, Wang, Zhou, and Qiu]{zhang2023speechgpt}
Dong Zhang, Shimin Li, Xin Zhang, Jun Zhan, Pengyu Wang, Yaqian Zhou, and Xipeng Qiu.
\newblock Speechgpt: Empowering large language models with intrinsic cross-modal conversational abilities, 2023{\natexlab{a}}.

\bibitem[Yang et~al.(2023{\natexlab{a}})Yang, Tian, Tan, Huang, Liu, Chang, Shi, Zhao, Bian, Wu, et~al.]{yang2023uniaudio}
Dongchao Yang, Jinchuan Tian, Xu~Tan, Rongjie Huang, Songxiang Liu, Xuankai Chang, Jiatong Shi, Sheng Zhao, Jiang Bian, Xixin Wu, et~al.
\newblock Uniaudio: An audio foundation model toward universal audio generation.
\newblock \emph{arXiv preprint arXiv:2310.00704}, 2023{\natexlab{a}}.

\bibitem[Chen et~al.(2023)Chen, Huang, Andrusenko, Hrinchuk, Puvvada, Li, Ghosh, Balam, and Ginsburg]{chen2023salm}
Zhehuai Chen, He~Huang, Andrei Andrusenko, Oleksii Hrinchuk, Krishna~C Puvvada, Jason Li, Subhankar Ghosh, Jagadeesh Balam, and Boris Ginsburg.
\newblock Salm: Speech-augmented language model with in-context learning for speech recognition and translation.
\newblock \emph{arXiv preprint arXiv:2310.09424}, 2023.

\bibitem[Kharitonov et~al.(2023)Kharitonov, Vincent, Borsos, Marinier, Girgin, Pietquin, Sharifi, Tagliasacchi, and Zeghidour]{10.1162/tacl_a_00618}
Eugene Kharitonov, Damien Vincent, Zalán Borsos, Raphaël Marinier, Sertan Girgin, Olivier Pietquin, Matt Sharifi, Marco Tagliasacchi, and Neil Zeghidour.
\newblock {Speak, Read and Prompt: High-Fidelity Text-to-Speech with Minimal Supervision}.
\newblock \emph{Transactions of the Association for Computational Linguistics}, 11:\penalty0 1703--1718, 12 2023.
\newblock ISSN 2307-387X.
\newblock \doi{10.1162/tacl_a_00618}.
\newblock URL \url{https://doi.org/10.1162/tacl\_a\_00618}.

\bibitem[Borsos et~al.(2023)Borsos, Marinier, Vincent, Kharitonov, Pietquin, Sharifi, Roblek, Teboul, Grangier, Tagliasacchi, et~al.]{borsos2023audiolm}
Zal{\'a}n Borsos, Rapha{\"e}l Marinier, Damien Vincent, Eugene Kharitonov, Olivier Pietquin, Matt Sharifi, Dominik Roblek, Olivier Teboul, David Grangier, Marco Tagliasacchi, et~al.
\newblock Audiolm: a language modeling approach to audio generation.
\newblock \emph{IEEE/ACM Transactions on Audio, Speech, and Language Processing}, 2023.

\bibitem[Hassid et~al.(2023)Hassid, Remez, Nguyen, Gat, Conneau, Kreuk, Copet, D{\'e}fossez, Synnaeve, Dupoux, Schwartz, and Adi]{hassid2023textually}
Michael Hassid, Tal Remez, Tu~Anh Nguyen, Itai Gat, Alexis Conneau, Felix Kreuk, Jade Copet, Alexandre D{\'e}fossez, Gabriel Synnaeve, Emmanuel Dupoux, Roy Schwartz, and Yossi Adi.
\newblock Textually pretrained speech language models.
\newblock In \emph{Thirty-seventh Conference on Neural Information Processing Systems}, 2023.
\newblock URL \url{https://openreview.net/forum?id=UlHueVjAKr}.

\bibitem[Lee et~al.(2022)Lee, Gong, Duquenne, Schwenk, Chen, Wang, Popuri, Adi, Pino, Gu, and Hsu]{lee-etal-2022-textless}
Ann Lee, Hongyu Gong, Paul-Ambroise Duquenne, Holger Schwenk, Peng-Jen Chen, Changhan Wang, Sravya Popuri, Yossi Adi, Juan Pino, Jiatao Gu, and Wei-Ning Hsu.
\newblock Textless speech-to-speech translation on real data.
\newblock In Marine Carpuat, Marie-Catherine de~Marneffe, and Ivan~Vladimir Meza~Ruiz, editors, \emph{Proceedings of the 2022 Conference of the North American Chapter of the Association for Computational Linguistics: Human Language Technologies}, pages 860--872, Seattle, United States, July 2022. Association for Computational Linguistics.
\newblock \doi{10.18653/v1/2022.naacl-main.63}.
\newblock URL \url{https://aclanthology.org/2022.naacl-main.63}.

\bibitem[Polyak et~al.(2021)Polyak, Adi, Copet, Kharitonov, Lakhotia, Hsu, rahman Mohamed, and Dupoux]{Polyak2021SpeechRF}
Adam Polyak, Yossi Adi, Jade Copet, Eugene Kharitonov, Kushal Lakhotia, Wei-Ning Hsu, Abdel rahman Mohamed, and Emmanuel Dupoux.
\newblock Speech resynthesis from discrete disentangled self-supervised representations.
\newblock In \emph{Interspeech}, 2021.
\newblock URL \url{https://api.semanticscholar.org/CorpusID:262491522}.

\bibitem[Communication et~al.(2023{\natexlab{a}})Communication, Barrault, Chung, Meglioli, Dale, Dong, Duppenthaler, Duquenne, Ellis, Elsahar, Haaheim, Hoffman, Hwang, Inaguma, Klaiber, Kulikov, Li, Licht, Maillard, Mavlyutov, Rakotoarison, Sadagopan, Ramakrishnan, Tran, Wenzek, Yang, Ye, Evtimov, Fernandez, Gao, Hansanti, Kalbassi, Kallet, Kozhevnikov, Gonzalez, Roman, Touret, Wong, Wood, Yu, Andrews, Balioglu, Chen, Costa-jussà, Elbayad, Gong, Guzmán, Heffernan, Jain, Kao, Lee, Ma, Mourachko, Peloquin, Pino, Popuri, Ropers, Saleem, Schwenk, Sun, Tomasello, Wang, Wang, Wang, and Williamson]{communication2023seamless}
Seamless Communication, Loïc Barrault, Yu-An Chung, Mariano~Coria Meglioli, David Dale, Ning Dong, Mark Duppenthaler, Paul-Ambroise Duquenne, Brian Ellis, Hady Elsahar, Justin Haaheim, John Hoffman, Min-Jae Hwang, Hirofumi Inaguma, Christopher Klaiber, Ilia Kulikov, Pengwei Li, Daniel Licht, Jean Maillard, Ruslan Mavlyutov, Alice Rakotoarison, Kaushik~Ram Sadagopan, Abinesh Ramakrishnan, Tuan Tran, Guillaume Wenzek, Yilin Yang, Ethan Ye, Ivan Evtimov, Pierre Fernandez, Cynthia Gao, Prangthip Hansanti, Elahe Kalbassi, Amanda Kallet, Artyom Kozhevnikov, Gabriel~Mejia Gonzalez, Robin~San Roman, Christophe Touret, Corinne Wong, Carleigh Wood, Bokai Yu, Pierre Andrews, Can Balioglu, Peng-Jen Chen, Marta~R. Costa-jussà, Maha Elbayad, Hongyu Gong, Francisco Guzmán, Kevin Heffernan, Somya Jain, Justine Kao, Ann Lee, Xutai Ma, Alex Mourachko, Benjamin Peloquin, Juan Pino, Sravya Popuri, Christophe Ropers, Safiyyah Saleem, Holger Schwenk, Anna Sun, Paden Tomasello, Changhan Wang, Jeff Wang, Skyler Wang, and Mary
  Williamson.
\newblock Seamless: Multilingual expressive and streaming speech translation, 2023{\natexlab{a}}.

\bibitem[Zeghidour et~al.(2021)Zeghidour, Luebs, Omran, Skoglund, and Tagliasacchi]{10.1109/TASLP.2021.3129994}
Neil Zeghidour, Alejandro Luebs, Ahmed Omran, Jan Skoglund, and Marco Tagliasacchi.
\newblock Soundstream: An end-to-end neural audio codec.
\newblock \emph{IEEE/ACM Trans. Audio, Speech and Lang. Proc.}, 30:\penalty0 495–507, nov 2021.
\newblock ISSN 2329-9290.
\newblock \doi{10.1109/TASLP.2021.3129994}.
\newblock URL \url{https://doi.org/10.1109/TASLP.2021.3129994}.

\bibitem[Défossez et~al.(2022)Défossez, Copet, Synnaeve, and Adi]{defossez2022highfi}
Alexandre Défossez, Jade Copet, Gabriel Synnaeve, and Yossi Adi.
\newblock High fidelity neural audio compression.
\newblock \emph{arXiv preprint arXiv:2210.13438}, 2022.

\bibitem[Kumar et~al.(2023)Kumar, Seetharaman, Luebs, Kumar, and Kumar]{kumar2023highfidelity}
Rithesh Kumar, Prem Seetharaman, Alejandro Luebs, Ishaan Kumar, and Kundan Kumar.
\newblock High-fidelity audio compression with improved {RVQGAN}.
\newblock In \emph{Thirty-seventh Conference on Neural Information Processing Systems}, 2023.
\newblock URL \url{https://openreview.net/forum?id=qjnl1QUnFA}.

\bibitem[Yang et~al.(2023{\natexlab{b}})Yang, Liu, Huang, Tian, Weng, and Zou]{yang2023hifi}
Dongchao Yang, Songxiang Liu, Rongjie Huang, Jinchuan Tian, Chao Weng, and Yuexian Zou.
\newblock Hifi-codec: Group-residual vector quantization for high fidelity audio codec.
\newblock \emph{arXiv preprint arXiv:2305.02765}, 2023{\natexlab{b}}.

\bibitem[Zhang et~al.(2024{\natexlab{a}})Zhang, Zhang, Li, Zhou, and Qiu]{zhang2024speechtokenizer}
Xin Zhang, Dong Zhang, Shimin Li, Yaqian Zhou, and Xipeng Qiu.
\newblock Speechtokenizer: Unified speech tokenizer for speech language models.
\newblock In \emph{The Twelfth International Conference on Learning Representations}, 2024{\natexlab{a}}.
\newblock URL \url{https://openreview.net/forum?id=AF9Q8Vip84}.

\bibitem[Vasuki and Vanathi(2006)]{1664069}
A.~Vasuki and P.T. Vanathi.
\newblock A review of vector quantization techniques.
\newblock \emph{IEEE Potentials}, 25\penalty0 (4):\penalty0 39--47, 2006.
\newblock \doi{10.1109/MP.2006.1664069}.

\bibitem[Kreuk et~al.(2023)Kreuk, Synnaeve, Polyak, Singer, D{\'e}fossez, Copet, Parikh, Taigman, and Adi]{kreuk2023audiogen}
Felix Kreuk, Gabriel Synnaeve, Adam Polyak, Uriel Singer, Alexandre D{\'e}fossez, Jade Copet, Devi Parikh, Yaniv Taigman, and Yossi Adi.
\newblock Audiogen: Textually guided audio generation.
\newblock In \emph{The Eleventh International Conference on Learning Representations}, 2023.
\newblock URL \url{https://openreview.net/forum?id=CYK7RfcOzQ4}.

\bibitem[Copet et~al.(2023)Copet, Kreuk, Gat, Remez, Kant, Synnaeve, Adi, and D{\'e}fossez]{copet2023simple}
Jade Copet, Felix Kreuk, Itai Gat, Tal Remez, David Kant, Gabriel Synnaeve, Yossi Adi, and Alexandre D{\'e}fossez.
\newblock Simple and controllable music generation.
\newblock In \emph{Thirty-seventh Conference on Neural Information Processing Systems}, 2023.
\newblock URL \url{https://openreview.net/forum?id=jtiQ26sCJi}.

\bibitem[Agostinelli et~al.(2023)Agostinelli, Denk, Borsos, Engel, Verzetti, Caillon, Huang, Jansen, Roberts, Tagliasacchi, Sharifi, Zeghidour, and Frank]{agostinelli2023musiclm}
Andrea Agostinelli, Timo~I. Denk, Zalán Borsos, Jesse Engel, Mauro Verzetti, Antoine Caillon, Qingqing Huang, Aren Jansen, Adam Roberts, Marco Tagliasacchi, Matt Sharifi, Neil Zeghidour, and Christian Frank.
\newblock Musiclm: Generating music from text, 2023.

\bibitem[Tang et~al.(2023)Tang, Yu, Sun, Chen, Tan, Li, Lu, Ma, and Zhang]{tang2023salmonn}
Changli Tang, Wenyi Yu, Guangzhi Sun, Xianzhao Chen, Tian Tan, Wei Li, Lu~Lu, Zejun Ma, and Chao Zhang.
\newblock Salmonn: Towards generic hearing abilities for large language models, 2023.

\bibitem[Wang et~al.(2023{\natexlab{b}})Wang, Liao, Huang, Lu, Wu, Liu, Zong, and Zhang]{wang2023blsp}
Chen Wang, Minpeng Liao, Zhongqiang Huang, Jinliang Lu, Junhong Wu, Yuchen Liu, Chengqing Zong, and Jiajun Zhang.
\newblock Blsp: Bootstrapping language-speech pre-training via behavior alignment of continuation writing, 2023{\natexlab{b}}.

\bibitem[Team et~al.(2023)Team, Anil, Borgeaud, Wu, Alayrac, Yu, Soricut, Schalkwyk, Dai, Hauth, et~al.]{team2023gemini}
Gemini Team, Rohan Anil, Sebastian Borgeaud, Yonghui Wu, Jean-Baptiste Alayrac, Jiahui Yu, Radu Soricut, Johan Schalkwyk, Andrew~M Dai, Anja Hauth, et~al.
\newblock Gemini: a family of highly capable multimodal models.
\newblock \emph{arXiv preprint arXiv:2312.11805}, 2023.

\bibitem[Hono et~al.(2023)Hono, Mitsuda, Zhao, Mitsui, Wakatsuki, and Sawada]{hono2023integration}
Yukiya Hono, Koh Mitsuda, Tianyu Zhao, Kentaro Mitsui, Toshiaki Wakatsuki, and Kei Sawada.
\newblock An integration of pre-trained speech and language models for end-to-end speech recognition.
\newblock \emph{arXiv preprint arXiv:2312.03668}, 2023.

\bibitem[Fathullah et~al.(2023{\natexlab{a}})Fathullah, Wu, Lakomkin, Jia, Shangguan, Li, Guo, Xiong, Mahadeokar, Kalinli, et~al.]{fathullah2023prompting}
Yassir Fathullah, Chunyang Wu, Egor Lakomkin, Junteng Jia, Yuan Shangguan, Ke~Li, Jinxi Guo, Wenhan Xiong, Jay Mahadeokar, Ozlem Kalinli, et~al.
\newblock Prompting large language models with speech recognition abilities.
\newblock \emph{arXiv preprint arXiv:2307.11795}, 2023{\natexlab{a}}.

\bibitem[Wang et~al.(2023{\natexlab{c}})Wang, Han, Shafran, Wu, Chiu, Cao, Wang, Chen, Zhang, Soltau, Rubenstein, Zilka, Yu, Meng, Pundak, Siddhartha, Schalkwyk, and Wu]{wang2023slm}
Mingqiu Wang, Wei Han, Izhak Shafran, Zelin Wu, Chung-Cheng Chiu, Yuan Cao, Yongqiang Wang, Nanxin Chen, Yu~Zhang, Hagen Soltau, Paul Rubenstein, Lukas Zilka, Dian Yu, Zhong Meng, Golan Pundak, Nikhil Siddhartha, Johan Schalkwyk, and Yonghui Wu.
\newblock Slm: Bridge the thin gap between speech and text foundation models, 2023{\natexlab{c}}.

\bibitem[Hu et~al.(2024)Hu, CHEN, Yang, Li, Zhang, Chen, and Chng]{hu2024large}
Yuchen Hu, CHEN CHEN, Chao-Han~Huck Yang, Ruizhe Li, Chao Zhang, Pin-Yu Chen, and EngSiong Chng.
\newblock Large language models are efficient learners of noise-robust speech recognition.
\newblock In \emph{The Twelfth International Conference on Learning Representations}, 2024.
\newblock URL \url{https://openreview.net/forum?id=ceATjGPTUD}.

\bibitem[Pan et~al.(2023)Pan, Wu, Gaur, Sivasankaran, Chen, Liu, and Li]{Pan2023COSMICDE}
Jing Pan, Jian Wu, Yashesh Gaur, Sunit Sivasankaran, Zhuo Chen, Shujie Liu, and Jinyu Li.
\newblock Cosmic: Data efficient instruction-tuning for speech in-context learning.
\newblock \emph{ArXiv}, abs/2311.02248, 2023.
\newblock URL \url{https://api.semanticscholar.org/CorpusID:265034072}.

\bibitem[Zhao et~al.(2023)Zhao, Jiang, Liu, Wang, and Wang]{zhao2023librisqa}
Zihan Zhao, Yiyang Jiang, Heyang Liu, Yanfeng Wang, and Yu~Wang.
\newblock Librisqa: Advancing free-form and open-ended spoken question answering with a novel dataset and framework.
\newblock \emph{arXiv preprint arXiv:2308.10390}, 2023.

\bibitem[Gong et~al.(2023)Gong, H, Luo, Karlinsky, and Glass]{gong_ltuas}
Yuan Gong, Liu~Alexander H, Hongyin Luo, Leonid Karlinsky, and James Glass.
\newblock Joint audio and speech understanding.
\newblock In \emph{2023 IEEE Automatic Speech Recognition and Understanding Workshop (ASRU)}, 2023.

\bibitem[Fathullah et~al.(2023{\natexlab{b}})Fathullah, Wu, Lakomkin, Jia, Shangguan, Mahadeokar, Kalinli, Fuegen, and Seltzer]{fathullah2023generalpurpose}
Yassir Fathullah, Chunyang Wu, Egor Lakomkin, Junteng Jia, Yuan Shangguan, Jay Mahadeokar, Ozlem Kalinli, Christian Fuegen, and Mike Seltzer.
\newblock Towards general-purpose speech abilities for large language models using unpaired data, 2023{\natexlab{b}}.

\bibitem[Zhang et~al.(2023{\natexlab{b}})Zhang, Zhou, Wang, Chen, Wu, Liu, Chen, Liu, Wang, Li, He, Zhao, and Wei]{zhang2023speak}
Ziqiang Zhang, Long Zhou, Chengyi Wang, Sanyuan Chen, Yu~Wu, Shujie Liu, Zhuo Chen, Yanqing Liu, Huaming Wang, Jinyu Li, Lei He, Sheng Zhao, and Furu Wei.
\newblock Speak foreign languages with your own voice: Cross-lingual neural codec language modeling.
\newblock arXiv, March 2023{\natexlab{b}}.
\newblock URL \url{https://www.microsoft.com/en-us/research/publication/speak-foreign-languages-with-your-own-voice-cross-lingual-neural-codec-language-modeling/}.

\bibitem[kim(2024)]{kim2024clamtts}
{CL}am-{TTS}: Improving neural codec language model for zero-shot text-to-speech.
\newblock In \emph{The Twelfth International Conference on Learning Representations}, 2024.
\newblock URL \url{https://openreview.net/forum?id=ofzeypWosV}.

\bibitem[Nguyen et~al.(2024)Nguyen, Muller, Yu, Costa-jussa, Elbayad, Popuri, Duquenne, Algayres, Mavlyutov, Gat, Synnaeve, Pino, Sagot, and Dupoux]{nguyen2024spiritlm}
Tu~Anh Nguyen, Benjamin Muller, Bokai Yu, Marta~R. Costa-jussa, Maha Elbayad, Sravya Popuri, Paul-Ambroise Duquenne, Robin Algayres, Ruslan Mavlyutov, Itai Gat, Gabriel Synnaeve, Juan Pino, Benoit Sagot, and Emmanuel Dupoux.
\newblock Spirit-lm: Interleaved spoken and written language model, 2024.

\bibitem[Nachmani et~al.(2023)Nachmani, Levkovitch, Hirsch, Salazar, Asawaroengchai, Mariooryad, Rivlin, Skerry-Ryan, and Ramanovich]{nachmani2023spoken}
Eliya Nachmani, Alon Levkovitch, Roy Hirsch, Julian Salazar, Chulayuth Asawaroengchai, Soroosh Mariooryad, Ehud Rivlin, RJ~Skerry-Ryan, and Michelle~Tadmor Ramanovich.
\newblock Spoken question answering and speech continuation using spectrogram-powered llm, 2023.

\bibitem[Wang et~al.(2023{\natexlab{d}})Wang, Zhou, Zhang, Wu, Liu, Gaur, Chen, Li, and Wei]{wang2023viola}
Tianrui Wang, Long Zhou, Ziqiang Zhang, Yu~Wu, Shujie Liu, Yashesh Gaur, Zhuo Chen, Jinyu Li, and Furu Wei.
\newblock Viola: Unified codec language models for speech recognition, synthesis, and translation.
\newblock \emph{arXiv preprint arXiv:2305.16107}, 2023{\natexlab{d}}.

\bibitem[Communication et~al.(2023{\natexlab{b}})Communication, Barrault, Chung, Meglioli, Dale, Dong, Duquenne, Elsahar, Gong, Heffernan, Hoffman, Klaiber, Li, Licht, Maillard, Rakotoarison, Sadagopan, Wenzek, Ye, Akula, Chen, Hachem, Ellis, Gonzalez, Haaheim, Hansanti, Howes, Huang, Hwang, Inaguma, Jain, Kalbassi, Kallet, Kulikov, Lam, Li, Ma, Mavlyutov, Peloquin, Ramadan, Ramakrishnan, Sun, Tran, Tran, Tufanov, Vogeti, Wood, Yang, Yu, Andrews, Balioglu, Costa-jussà, Celebi, Elbayad, Gao, Guzmán, Kao, Lee, Mourachko, Pino, Popuri, Ropers, Saleem, Schwenk, Tomasello, Wang, Wang, and Wang]{communication2023seamlessm4t}
Seamless Communication, Loïc Barrault, Yu-An Chung, Mariano~Cora Meglioli, David Dale, Ning Dong, Paul-Ambroise Duquenne, Hady Elsahar, Hongyu Gong, Kevin Heffernan, John Hoffman, Christopher Klaiber, Pengwei Li, Daniel Licht, Jean Maillard, Alice Rakotoarison, Kaushik~Ram Sadagopan, Guillaume Wenzek, Ethan Ye, Bapi Akula, Peng-Jen Chen, Naji~El Hachem, Brian Ellis, Gabriel~Mejia Gonzalez, Justin Haaheim, Prangthip Hansanti, Russ Howes, Bernie Huang, Min-Jae Hwang, Hirofumi Inaguma, Somya Jain, Elahe Kalbassi, Amanda Kallet, Ilia Kulikov, Janice Lam, Daniel Li, Xutai Ma, Ruslan Mavlyutov, Benjamin Peloquin, Mohamed Ramadan, Abinesh Ramakrishnan, Anna Sun, Kevin Tran, Tuan Tran, Igor Tufanov, Vish Vogeti, Carleigh Wood, Yilin Yang, Bokai Yu, Pierre Andrews, Can Balioglu, Marta~R. Costa-jussà, Onur Celebi, Maha Elbayad, Cynthia Gao, Francisco Guzmán, Justine Kao, Ann Lee, Alexandre Mourachko, Juan Pino, Sravya Popuri, Christophe Ropers, Safiyyah Saleem, Holger Schwenk, Paden Tomasello, Changhan Wang, Jeff
  Wang, and Skyler Wang.
\newblock Seamlessm4t: Massively multilingual \& multimodal machine translation, 2023{\natexlab{b}}.

\bibitem[qian Dong et~al.(2024)qian Dong, Huang, Tian, Xu, Ko, yunlong zhao, Feng, Li, Wang, Cheng, Yue, Bai, Chen, Lu, MA, Wang, Wang, and Wang]{dong2024polyvoice}
Qian qian Dong, Zhiying Huang, Qiao Tian, Chen Xu, Tom Ko, yunlong zhao, Siyuan Feng, Tang Li, Kexin Wang, Xuxin Cheng, Fengpeng Yue, Ye~Bai, Xi~Chen, Lu~Lu, Zejun MA, Yuping Wang, Mingxuan Wang, and Yuxuan Wang.
\newblock Polyvoice: Language models for speech to speech translation.
\newblock In \emph{The Twelfth International Conference on Learning Representations}, 2024.
\newblock URL \url{https://openreview.net/forum?id=hCrFG9cyuC}.

\bibitem[Zhan et~al.(2024)Zhan, Dai, Ye, Zhou, Zhang, Liu, Zhang, Yuan, Zhang, Li, et~al.]{zhan2024anygpt}
Jun Zhan, Junqi Dai, Jiasheng Ye, Yunhua Zhou, Dong Zhang, Zhigeng Liu, Xin Zhang, Ruibin Yuan, Ge~Zhang, Linyang Li, et~al.
\newblock Anygpt: Unified multimodal llm with discrete sequence modeling.
\newblock \emph{arXiv preprint arXiv:2402.12226}, 2024.

\bibitem[Lin et~al.(2024)Lin, Chiang, and yi~Lee]{lin2024advancing}
Guan-Ting Lin, Cheng-Han Chiang, and Hung yi~Lee.
\newblock Advancing large language models to capture varied speaking styles and respond properly in spoken conversations, 2024.

\bibitem[Bastianelli et~al.(2020)Bastianelli, Vanzo, Swietojanski, and Rieser]{bastianelli-etal-2020-slurp}
Emanuele Bastianelli, Andrea Vanzo, Pawel Swietojanski, and Verena Rieser.
\newblock {SLURP}: A spoken language understanding resource package.
\newblock In Bonnie Webber, Trevor Cohn, Yulan He, and Yang Liu, editors, \emph{Proceedings of the 2020 Conference on Empirical Methods in Natural Language Processing (EMNLP)}, pages 7252--7262, Online, November 2020. Association for Computational Linguistics.
\newblock \doi{10.18653/v1/2020.emnlp-main.588}.
\newblock URL \url{https://aclanthology.org/2020.emnlp-main.588}.

\bibitem[Babu et~al.(2022)Babu, Wang, Tjandra, Lakhotia, Xu, Goyal, Singh, {von Platen}, Saraf, Pino, Baevski, Conneau, and Auli]{babu22_interspeech}
Arun Babu, Changhan Wang, Andros Tjandra, Kushal Lakhotia, Qiantong Xu, Naman Goyal, Kritika Singh, Patrick {von Platen}, Yatharth Saraf, Juan Pino, Alexei Baevski, Alexis Conneau, and Michael Auli.
\newblock {XLS-R: Self-supervised Cross-lingual Speech Representation Learning at Scale}.
\newblock In \emph{Proc. Interspeech 2022}, pages 2278--2282, 2022.
\newblock \doi{10.21437/Interspeech.2022-143}.

\bibitem[Cao et~al.(2014)Cao, Cooper, Keutmann, Gur, Nenkova, and Verma]{6849440}
Houwei Cao, David~G. Cooper, Michael~K. Keutmann, Ruben~C. Gur, Ani Nenkova, and Ragini Verma.
\newblock Crema-d: Crowd-sourced emotional multimodal actors dataset.
\newblock \emph{IEEE Transactions on Affective Computing}, 5\penalty0 (4):\penalty0 377--390, 2014.
\newblock \doi{10.1109/TAFFC.2014.2336244}.

\bibitem[Le et~al.(2023)Le, Vyas, Shi, Karrer, Sari, Moritz, Williamson, Manohar, Adi, Mahadeokar, and Hsu]{le2023voicebox}
Matthew Le, Apoorv Vyas, Bowen Shi, Brian Karrer, Leda Sari, Rashel Moritz, Mary Williamson, Vimal Manohar, Yossi Adi, Jay Mahadeokar, and Wei-Ning Hsu.
\newblock Voicebox: Text-guided multilingual universal speech generation at scale.
\newblock In \emph{Thirty-seventh Conference on Neural Information Processing Systems}, 2023.
\newblock URL \url{https://openreview.net/forum?id=gzCS252hCO}.

\bibitem[Jiang et~al.(2023)Jiang, Sablayrolles, Mensch, Bamford, Chaplot, de~las Casas, Bressand, Lengyel, Lample, Saulnier, Lavaud, Lachaux, Stock, Scao, Lavril, Wang, Lacroix, and Sayed]{jiang2023mistral}
Albert~Q. Jiang, Alexandre Sablayrolles, Arthur Mensch, Chris Bamford, Devendra~Singh Chaplot, Diego de~las Casas, Florian Bressand, Gianna Lengyel, Guillaume Lample, Lucile Saulnier, Lélio~Renard Lavaud, Marie-Anne Lachaux, Pierre Stock, Teven~Le Scao, Thibaut Lavril, Thomas Wang, Timothée Lacroix, and William~El Sayed.
\newblock Mistral 7b, 2023.

\bibitem[McAuliffe et~al.(2017)McAuliffe, Socolof, Mihuc, Wagner, and Sonderegger]{mcauliffe17_interspeech}
Michael McAuliffe, Michaela Socolof, Sarah Mihuc, Michael Wagner, and Morgan Sonderegger.
\newblock {Montreal Forced Aligner: Trainable Text-Speech Alignment Using Kaldi}.
\newblock In \emph{Proc. Interspeech 2017}, pages 498--502, 2017.
\newblock \doi{10.21437/Interspeech.2017-1386}.

\bibitem[Nye et~al.(2022)Nye, Andreassen, Gur-Ari, Michalewski, Austin, Bieber, Dohan, Lewkowycz, Bosma, Luan, et~al.]{nye2022show}
Maxwell Nye, Anders~Johan Andreassen, Guy Gur-Ari, Henryk Michalewski, Jacob Austin, David Bieber, David Dohan, Aitor Lewkowycz, Maarten Bosma, David Luan, et~al.
\newblock Show your work: Scratchpads for intermediate computation with language models.
\newblock In \emph{Deep Learning for Code Workshop}, 2022.

\bibitem[Lee et~al.(2023)Lee, Park, and Kim]{10095751}
Keon Lee, Kyumin Park, and Daeyoung Kim.
\newblock Dailytalk: Spoken dialogue dataset for conversational text-to-speech.
\newblock In \emph{ICASSP 2023 - 2023 IEEE International Conference on Acoustics, Speech and Signal Processing (ICASSP)}, pages 1--5, 2023.
\newblock \doi{10.1109/ICASSP49357.2023.10095751}.

\bibitem[Pratap et~al.(2020)Pratap, Xu, Sriram, Synnaeve, and Collobert]{pratap20_interspeech}
Vineel Pratap, Qiantong Xu, Anuroop Sriram, Gabriel Synnaeve, and Ronan Collobert.
\newblock {MLS: A Large-Scale Multilingual Dataset for Speech Research}.
\newblock In \emph{Proc. Interspeech 2020}, pages 2757--2761, 2020.
\newblock \doi{10.21437/Interspeech.2020-2826}.

\bibitem[Chen et~al.(2021)Chen, Chai, Wang, Du, Zhang, Weng, Su, Povey, Trmal, Zhang, Jin, Khudanpur, Watanabe, Zhao, Zou, Li, Yao, Wang, Wang, You, and Yan]{GigaSpeech2021}
Guoguo Chen, Shuzhou Chai, Guanbo Wang, Jiayu Du, Wei-Qiang Zhang, Chao Weng, Dan Su, Daniel Povey, Jan Trmal, Junbo Zhang, Mingjie Jin, Sanjeev Khudanpur, Shinji Watanabe, Shuaijiang Zhao, Wei Zou, Xiangang Li, Xuchen Yao, Yongqing Wang, Yujun Wang, Zhao You, and Zhiyong Yan.
\newblock Gigaspeech: An evolving, multi-domain asr corpus with 10,000 hours of transcribed audio.
\newblock In \emph{Proc. Interspeech 2021}, 2021.

\bibitem[Kingma and Ba(2015)]{KingBa15}
Diederik Kingma and Jimmy Ba.
\newblock Adam: A method for stochastic optimization.
\newblock In \emph{International Conference on Learning Representations (ICLR)}, San Diega, CA, USA, 2015.

\bibitem[gil Lee et~al.(2023)gil Lee, Ping, Ginsburg, Catanzaro, and Yoon]{lee2023bigvgan}
Sang gil Lee, Wei Ping, Boris Ginsburg, Bryan Catanzaro, and Sungroh Yoon.
\newblock Big{VGAN}: A universal neural vocoder with large-scale training.
\newblock In \emph{The Eleventh International Conference on Learning Representations}, 2023.
\newblock URL \url{https://openreview.net/forum?id=iTtGCMDEzS_}.

\bibitem[Galvez et~al.(2021)Galvez, Diamos, Torres, Cer{\'o}n, Achorn, Gopi, Kanter, Lam, Mazumder, and Reddi]{galvez2021the}
Daniel Galvez, Greg Diamos, Juan Manuel~Ciro Torres, Juan~Felipe Cer{\'o}n, Keith Achorn, Anjali Gopi, David Kanter, Max Lam, Mark Mazumder, and Vijay~Janapa Reddi.
\newblock The people{\textquoteright}s speech: A large-scale diverse english speech recognition dataset for commercial usage.
\newblock In \emph{Thirty-fifth Conference on Neural Information Processing Systems Datasets and Benchmarks Track (Round 1)}, 2021.
\newblock URL \url{https://openreview.net/forum?id=R8CwidgJ0yT}.

\bibitem[Ardila et~al.(2020)Ardila, Branson, Davis, Kohler, Meyer, Henretty, Morais, Saunders, Tyers, and Weber]{ardila-etal-2020-common}
Rosana Ardila, Megan Branson, Kelly Davis, Michael Kohler, Josh Meyer, Michael Henretty, Reuben Morais, Lindsay Saunders, Francis Tyers, and Gregor Weber.
\newblock Common voice: A massively-multilingual speech corpus.
\newblock In Nicoletta Calzolari, Fr{\'e}d{\'e}ric B{\'e}chet, Philippe Blache, Khalid Choukri, Christopher Cieri, Thierry Declerck, Sara Goggi, Hitoshi Isahara, Bente Maegaard, Joseph Mariani, H{\'e}l{\`e}ne Mazo, Asuncion Moreno, Jan Odijk, and Stelios Piperidis, editors, \emph{Proceedings of the Twelfth Language Resources and Evaluation Conference}, pages 4218--4222, Marseille, France, May 2020. European Language Resources Association.
\newblock ISBN 979-10-95546-34-4.
\newblock URL \url{https://aclanthology.org/2020.lrec-1.520}.

\bibitem[Wang et~al.(2021)Wang, Riviere, Lee, Wu, Talnikar, Haziza, Williamson, Pino, and Dupoux]{wang-etal-2021-voxpopuli}
Changhan Wang, Morgane Riviere, Ann Lee, Anne Wu, Chaitanya Talnikar, Daniel Haziza, Mary Williamson, Juan Pino, and Emmanuel Dupoux.
\newblock {V}ox{P}opuli: A large-scale multilingual speech corpus for representation learning, semi-supervised learning and interpretation.
\newblock In Chengqing Zong, Fei Xia, Wenjie Li, and Roberto Navigli, editors, \emph{Proceedings of the 59th Annual Meeting of the Association for Computational Linguistics and the 11th International Joint Conference on Natural Language Processing (Volume 1: Long Papers)}, pages 993--1003, Online, August 2021. Association for Computational Linguistics.
\newblock \doi{10.18653/v1/2021.acl-long.80}.
\newblock URL \url{https://aclanthology.org/2021.acl-long.80}.

\bibitem[Radford et~al.(2023)Radford, Kim, Xu, Brockman, Mcleavey, and Sutskever]{pmlr-v202-radford23a}
Alec Radford, Jong~Wook Kim, Tao Xu, Greg Brockman, Christine Mcleavey, and Ilya Sutskever.
\newblock Robust speech recognition via large-scale weak supervision.
\newblock In Andreas Krause, Emma Brunskill, Kyunghyun Cho, Barbara Engelhardt, Sivan Sabato, and Jonathan Scarlett, editors, \emph{Proceedings of the 40th International Conference on Machine Learning}, volume 202 of \emph{Proceedings of Machine Learning Research}, pages 28492--28518. PMLR, 23--29 Jul 2023.
\newblock URL \url{https://proceedings.mlr.press/v202/radford23a.html}.

\bibitem[Sai et~al.(2020)Sai, Mohankumar, Arora, and Khapra]{10.1162/tacl_a_00347}
Ananya~B. Sai, Akash~Kumar Mohankumar, Siddhartha Arora, and Mitesh~M. Khapra.
\newblock {Improving Dialog Evaluation with a Multi-reference Adversarial Dataset and Large Scale Pretraining}.
\newblock \emph{Transactions of the Association for Computational Linguistics}, 8:\penalty0 810--827, 12 2020.
\newblock ISSN 2307-387X.
\newblock \doi{10.1162/tacl_a_00347}.
\newblock URL \url{https://doi.org/10.1162/tacl\_a\_00347}.

\bibitem[Zhang et~al.(2020)Zhang, Sun, Galley, Chen, Brockett, Gao, Gao, Liu, and Dolan]{zhang2019dialogpt}
Yizhe Zhang, Siqi Sun, Michel Galley, Yen-Chun Chen, Chris Brockett, Xiang Gao, Jianfeng Gao, Jingjing Liu, and Bill Dolan.
\newblock Dialogpt: Large-scale generative pre-training for conversational response generation.
\newblock In \emph{ACL, system demonstration}, 2020.

\bibitem[Zheng et~al.(2023)Zheng, Chiang, Sheng, Zhuang, Wu, Zhuang, Lin, Li, Li, Xing, Zhang, Gonzalez, and Stoica]{zheng2023judging}
Lianmin Zheng, Wei-Lin Chiang, Ying Sheng, Siyuan Zhuang, Zhanghao Wu, Yonghao Zhuang, Zi~Lin, Zhuohan Li, Dacheng Li, Eric Xing, Hao Zhang, Joseph~E. Gonzalez, and Ion Stoica.
\newblock Judging {LLM}-as-a-judge with {MT}-bench and chatbot arena.
\newblock In \emph{Thirty-seventh Conference on Neural Information Processing Systems Datasets and Benchmarks Track}, 2023.
\newblock URL \url{https://openreview.net/forum?id=uccHPGDlao}.

\bibitem[Panayotov et~al.(2015)Panayotov, Chen, Povey, and Khudanpur]{7178964}
Vassil Panayotov, Guoguo Chen, Daniel Povey, and Sanjeev Khudanpur.
\newblock Librispeech: An asr corpus based on public domain audio books.
\newblock In \emph{2015 IEEE International Conference on Acoustics, Speech and Signal Processing (ICASSP)}, pages 5206--5210, 2015.
\newblock \doi{10.1109/ICASSP.2015.7178964}.

\bibitem[Bannò et~al.(2024)Bannò, Ma, Qian, Knill, and Gales]{10446782}
Stefano Bannò, Rao Ma, Mengjie Qian, Kate~M. Knill, and Mark J.~F. Gales.
\newblock Towards end-to-end spoken grammatical error correction.
\newblock In \emph{ICASSP 2024 - 2024 IEEE International Conference on Acoustics, Speech and Signal Processing (ICASSP)}, pages 10791--10795, 2024.
\newblock \doi{10.1109/ICASSP48485.2024.10446782}.

\bibitem[Zhang et~al.(2024{\natexlab{b}})Zhang, Zhang, Liu, Ye, Zhou, Lin, and Dai]{10446635}
Weitai Zhang, Hanyi Zhang, Chenxuan Liu, Zhongyi Ye, Xinyuan Zhou, Chao Lin, and Lirong Dai.
\newblock Pre-trained acoustic-and-textual modeling for end-to-end speech-to-text translation.
\newblock In \emph{ICASSP 2024 - 2024 IEEE International Conference on Acoustics, Speech and Signal Processing (ICASSP)}, pages 11451--11455, 2024{\natexlab{b}}.
\newblock \doi{10.1109/ICASSP48485.2024.10446635}.

\bibitem[Wang et~al.(2023{\natexlab{e}})Wang, Li, Guo, Qiao, Li, Shang, Wei, Tao, Zhang, and Yang]{wang23ga_interspeech}
Minghan Wang, Yinglu Li, Jiaxin Guo, Xiaosong Qiao, Zongyao Li, Hengchao Shang, Daimeng Wei, Shimin Tao, Min Zhang, and Hao Yang.
\newblock {WhiSLU: End-to-End Spoken Language Understanding with Whisper}.
\newblock In \emph{Proc. INTERSPEECH 2023}, pages 770--774, 2023{\natexlab{e}}.
\newblock \doi{10.21437/Interspeech.2023-1505}.

\bibitem[Xue et~al.(2022)Xue, Wang, Li, Post, and Gaur]{Xue2022LargeScaleSE}
Jian Xue, Peidong Wang, Jinyu Li, Matt Post, and Yashesh Gaur.
\newblock Large-scale streaming end-to-end speech translation with neural transducers.
\newblock In \emph{Interspeech}, 2022.
\newblock URL \url{https://api.semanticscholar.org/CorpusID:248118691}.

\bibitem[Nguyen et~al.(2023{\natexlab{b}})Nguyen, Hsu, D'Avirro, Shi, Gat, Fazel-Zarani, Remez, Copet, Synnaeve, Hassid, Kreuk, Adi, and Dupoux]{nguyen23_interspeech}
Tu~Anh Nguyen, Wei-Ning Hsu, Antony D'Avirro, Bowen Shi, Itai Gat, Maryam Fazel-Zarani, Tal Remez, Jade Copet, Gabriel Synnaeve, Michael Hassid, Felix Kreuk, Yossi Adi, and Emmanuel Dupoux.
\newblock {Expresso: A Benchmark and Analysis of Discrete Expressive Speech Resynthesis}.
\newblock In \emph{Proc. INTERSPEECH 2023}, pages 4823--4827, 2023{\natexlab{b}}.
\newblock \doi{10.21437/Interspeech.2023-1905}.

\bibitem[Cieri et~al.(2004)Cieri, Miller, and Walker]{cieri-etal-2004-fisher}
Christopher Cieri, David Miller, and Kevin Walker.
\newblock The fisher corpus: a resource for the next generations of speech-to-text.
\newblock In Maria~Teresa Lino, Maria~Francisca Xavier, F{\'a}tima Ferreira, Rute Costa, and Raquel Silva, editors, \emph{Proceedings of the Fourth International Conference on Language Resources and Evaluation ({LREC}{'}04)}, Lisbon, Portugal, May 2004. European Language Resources Association (ELRA).
\newblock URL \url{http://www.lrec-conf.org/proceedings/lrec2004/pdf/767.pdf}.

\bibitem[Lipman et~al.(2023)Lipman, Chen, Ben-Hamu, Nickel, and Le]{lipman2023flow}
Yaron Lipman, Ricky T.~Q. Chen, Heli Ben-Hamu, Maximilian Nickel, and Matthew Le.
\newblock Flow matching for generative modeling.
\newblock In \emph{The Eleventh International Conference on Learning Representations}, 2023.
\newblock URL \url{https://openreview.net/forum?id=PqvMRDCJT9t}.

\bibitem[Hu et~al.(2022)Hu, yelong shen, Wallis, Allen-Zhu, Li, Wang, Wang, and Chen]{hu2022lora}
Edward~J Hu, yelong shen, Phillip Wallis, Zeyuan Allen-Zhu, Yuanzhi Li, Shean Wang, Lu~Wang, and Weizhu Chen.
\newblock Lo{RA}: Low-rank adaptation of large language models.
\newblock In \emph{International Conference on Learning Representations}, 2022.
\newblock URL \url{https://openreview.net/forum?id=nZeVKeeFYf9}.

\end{thebibliography}
\bibliographystyle{unsrtnat}

\appendix

\newpage

\section{Appendix}

\subsection{Audio Samples}
\label{appendix:audio samples}
We have included various audio samples from our experiments on our demo page.\footnote{Our demo is available at \href{https://unifiedsdm.github.io/}{https://unifiedsdm.github.io/}.} Furthermore, to demonstrate the applicability and potential of our model, we have added several samples of USDM fine-tuned on the Expresso dataset \cite{nguyen23_interspeech}, which contains emotionally rich spoken dialog data, and the Fisher dataset \cite{cieri-etal-2004-fisher}, a telephony conversation dataset between two speakers recorded at 8,000Hz, totaling approximately 1,960 hours from 11,971 speakers.

The Expresso dataset comprises 41 hours of emotionally expressive speech data from 4 speakers. Of this, 11 hours consist of simple reading styles, while 30 hours are improvised dialogs between two speakers. We use this dialog data to train our USDM, noting that these improvised dialogs lack corresponding transcripts. Therefore, we create transcripts using \textit{whisper-large-v3} and utilize these to train the USDM. We observe that the trained model often failed in the ASR and TTS parts due to numerous inaccuracies in the transcripts generated by the ASR model, thus not using this data in the main experiments of our paper. Instead, we provide selected samples from the model trained on this data on our demo page.

In addition, we extend the spoken dialog templates used in training USDM to multi-turn scenarios to show the capability of USDM for modeling multi-turn dialogs. We explore this possibility by fine-tuning our speech-text model with Fisher \cite{cieri-etal-2004-fisher}. We split the train and test sets with no overlapping speakers to show the possibility of unseen speakers’ spoken dialog modeling. We have included samples of USDM's generated responses for multi-turn dialogs with unseen speakers of Fisher on our demo page. These samples demonstrate the potential of USDM for multi-turn spoken dialog modeling.

\subsection{Broader Impacts}
\label{appendix:impacts}
This paper proposes a spoken dialog model designed to generate spoken responses to the speech inputs. Similar to the field of natural language processing, the dialog model might exhibit biases in its outputs, which stem from the training dataset. Such biases could unintentionally lead to the generation of synthesized voices that are biased. Furthermore, there are ethical considerations regarding the possible misuse of high-quality speech synthesis models, such as in voice phishing scams.

Despite these concerns, research into spoken dialog models can yield several positive effects. Unlike text, voice interactions are capable of conveying non-verbal information, allowing us to build conversational agents that consider users' emotions, which are challenging to capture with text-based dialog models. Furthermore, by introducing spoken language as an additional means of communication, we offer an alternative to text-based chatbots for individuals facing difficulties in reading and writing. Similar to text-based chatbots, which have rapidly evolved and brought convenience to daily life, we expect that, with careful consideration of ethical issues, research and development in spoken dialog models will significantly benefit everyday life and a wide range of industries.

\subsection{Additional Details for Our Approach}
\subsubsection{Emotional Cues in Acoustic Units}
\label{appendix:emotional_cue}

In Section \ref{approach:prosody_unit}, we demonstrate through two experiments that the acoustic units extracted from XLS-R \cite{babu22_interspeech} contain paralinguistic features. The first experiment utilize CREMA-D \cite{6849440}, to perform a unit-to-emotion recognition task, to check whether the unit sequence contains information beyond content. Additionally, we train a unit-to-speech reconstruction module and use it to compare the original audio with the audio reconstructed from the extracted units, showing that the unit contains information about pitch variations, as shown in Figure \ref{fig2:f0_contour} and several samples on our demo page. Further details about the trained unit-to-speech reconstruction module are provided in Section \ref{appendix:voicebox}.

For emotion recognition, we train a 3-layer transformer-based emotion classifier using cross-entropy loss. We utilize the CREMA-D dataset, consisting of 7,442 audios from 91 actors, categorized into six emotions: Anger, Disgust, Fear, Happy, Neutral, and Sad. We split the data into training, validation, and testing sets in a ratio of $70\%$, $15\%$, and $15\%$, respectively, ensuring an equal number of samples for each emotion in both the validation and test sets.



\subsubsection{Templates for Fine-tuning}
\label{appendix:fine-tuning template}

\begin{figure}
    \centering
    \includegraphics[width=0.9\linewidth]{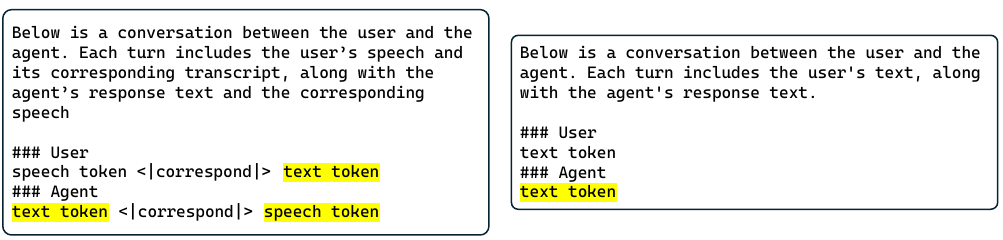} 
    \caption{Fine-tuning template for single-turn spoken dialog modeling. Left is the template used for training spoken dialog models (USDM, From Scratch), while the right is the template for training a text dialog model (Cascaded).}
    \label{fig:finetuning-template}
\end{figure}

Figure \ref{fig:finetuning-template} illustrates the template of single-turn spoken dialog and text dialog we used for fine-tuning. We perform loss calculations only on the highlighted part. USDM refers to our model, and From Scratch and Cascaded are the baselines we used as described in Section \ref{experiments:details}.

\begin{table}[]
\caption{Models for each component of the USDM and the baselines.}
\label{table:component}
\small
\begin{center}
\begin{tabular}{l|ccccc}
\toprule
\multicolumn{1}{c|}{\textbf{Model}} & \textbf{ASR Model} & \textbf{Speech Encoder} & \textbf{LLM} & \textbf{Speech Decoder}                                           & \textbf{TTS Model}                                           \\ \midrule
USDM                                & $-$                & XLS-R                   & Mistral-7B                & \begin{tabular}[c]{@{}c@{}}unit-Voicebox\\ + BigVGAN\end{tabular} & $-$                                                          \\ \midrule
From Scratch                        & $-$                & XLS-R                   & Mistral-7B                & \begin{tabular}[c]{@{}c@{}}unit-Voicebox\\ + BigVGAN\end{tabular} & $-$                                                          \\ \midrule
SpeechGPT                           & $-$                & mHuBERT                 & Llama-7B                  & unit-HiFi-GAN                                                     & $-$                                                          \\ \midrule
Cascaded                            & \textit{whisper-large-v3}   & $-$                     & Mistral-7B                & $-$                                                               & \begin{tabular}[c]{@{}c@{}}Voicebox\\ + BigVGAN\end{tabular}         \\
\bottomrule
\end{tabular}
\end{center}
\end{table}

\subsubsection{Voicebox}
\label{appendix:voicebox}
Voicebox \cite{le2023voicebox} is a Flow-Matching-based zero-shot TTS model \cite{lipman2023flow} that generates a mel-spectrogram from input text and reference speech. During training, \citet{le2023voicebox} utilize the Montreal Forced Aligner (MFA) \cite{mcauliffe17_interspeech} to extract the alignment between the phoneme sequence and the mel-spectrogram. This alignment is utilized to calculate the duration for each phoneme, allowing \citet{le2023voicebox} to expand the phoneme sequence to match the length of the mel-spectrogram. Voicebox is trained to produce a mel-spectrogram from an expanded phoneme sequence of equivalent length. \citet{le2023voicebox} additionally train a duration predictor to predict the duration of each phoneme in the sequence, which is used for inference.

We train three variants of the Voicebox for different purposes, all using the English subset of the Multilingual LibriSpeech \cite{pratap20_interspeech} and GigaSpeech \cite{GigaSpeech2021}, totaling 54k hours of ASR data. We perform inference using a total of 50 timesteps with a classifier-free guidance scale of 1. We train the model to produce mel-spectrograms of approximately 86Hz, with the same configuration as the official 22,050Hz checkpoint of BigVGAN \cite{lee2023bigvgan}. The output mel-spectrograms of all Voicebox models are converted into 22,050Hz speech through BigVGAN. This section will detail each of these 3 cases.

\textbf{Variant for Analyzing Non-Verbal Cues in Acoustic Unit} As described in Section \ref{approach:prosody_unit}, we train a unit-to-speech reconstruction model to investigate the non-verbal information contained within the unit. This model differs from Voicebox in that it utilizes unit sequence instead of text input and does not use reference speech during training and inference. We do not train a separate duration predictor; instead, we upsample the 50Hz unit sequence to 86Hz to match the length of the mel-spectrogram. Note that this variant is used only for analyzing non-verbal information.

\textbf{Unit-Voicebox for USDM} This Unit-Voicebox is a speech decoder used to restore speech from the output unit sequence of our model and From Scratch. It is trained in the same manner as the aforementioned variant but utilizes reference speech during training and inference to enable zero-shot speech reconstruction. To consistently respond in the same voice in multi-turn dialog scenarios, we use the adaptive TTS model, Voicebox as our speech reconstruction model, and leverage the spoken response of the preceding turn as the reference speech.

\textbf{Voicebox for Cascaded} This is the TTS model for one of our baseline models, the Cascaded model. We set the TTS model of the Cascaded model to Voicebox, training it with the same data and method as unit-Voicebox for a fair comparison. Unlike the unit-Voicebox, which does not require a duration predictor, we train a separate feed-forward duration predictor following \citet{le2023voicebox}.

\subsection{Additional Details for Evaluation}
\subsubsection{Models, Datasets, Training Details}
\label{appendix:models_datasets}
We list the models we used for each component of our model and the baselines in Table \ref{table:component}. In Table \ref{table:component}, `LLM' refers to the language model used prior to performing speech-text pretraining and/or fine-tuning.
Additionally, we list the licenses of the datasets used in Table \ref{table:license}, and include links to the open-source implementations, checkpoints, and packages we use in Table \ref{table:link}.

We utilize the DailyTalk dataset to evaluate the performance of USDM. We follow the train/test split of \citet{10095751} and preprocess the data for single-turn spoken dialog. As a result, we obtain a total of 20,117 training samples and 1,058 test samples.

\begin{table}[]
\caption{License of each dataset we used for acoustic unit investigation, pretraining, and fine-tuning.}
\label{table:license}
\small
\begin{center}
\begin{tabular}{l|cccc}
\toprule
\textbf{Dataset}   & \textbf{Unit Analysis}      & \textbf{Pretraining} & \textbf{Fine-tuning}  & \textbf{License}               \\ \midrule
CREMA-D \cite{6849440} & \cmark & \xmark & \xmark & Open Database License    
\\ \midrule
Multilingual LibriSpeech \cite{pratap20_interspeech} & \cmark & \cmark & \xmark & CC-BY-4.0                                                                                               \\
People's Speech \cite{galvez2021the} & \xmark         & \cmark & \xmark & CC-BY-SA                                                                                                \\
GigaSpeech \cite{GigaSpeech2021}   & \cmark            & \cmark & \xmark & Apache-2.0                                                                                              \\
Common Voice 15.0 \cite{ardila-etal-2020-common}  & \xmark      & \cmark & \xmark & CC-0                                                                                                    \\
Voxpopuli \cite{wang-etal-2021-voxpopuli}    & \xmark            & \cmark & \xmark & CC-0                                                                                                    \\ \midrule
DailyTalk \cite{10095751}     & \xmark           & \xmark & \cmark & CC-BY-SA 4.0                                                                                            \\
Expresso \cite{nguyen23_interspeech}      & \xmark             & \xmark & \cmark & CC BY-NC 4.0 \\
Fisher \cite{cieri-etal-2004-fisher}      & \xmark             & \xmark & \cmark & LDC User Agreement \\
\bottomrule
\end{tabular}
\end{center}
\end{table}

\begin{table}[]
\caption{Links to the model implementations, checkpoints, and libraries used.}
\label{table:link}
\small
\begin{center}
\begin{tabular}{l|l}
\toprule
                                          & \textbf{Link} \\ \midrule
\begin{tabular}[c]{@{}l@{}}XLS-R-based \\ Unit Extractor \cite{babu22_interspeech} \end{tabular}                 & \href{https://github.com/facebookresearch/seamless_communication}{https://github.com/facebookresearch/seamless\_communication}              \\ \midrule
Mistral-7B \cite{jiang2023mistral}                               & \href{https://huggingface.co/mistralai/Mistral-7B-v0.1}{https://huggingface.co/mistralai/Mistral-7B-v0.1}              \\ \midrule
SpeechGPT \cite{zhang2023speechgpt}                                & \href{https://github.com/0nutation/SpeechGPT/tree/main/speechgpt}{https://github.com/0nutation/SpeechGPT/tree/main/speechgpt}              \\ \midrule
\textit{whisper-large-v3} \cite{pmlr-v202-radford23a}                         & \href{https://huggingface.co/openai/whisper-large-v3}{https://huggingface.co/openai/whisper-large-v3}              \\ \midrule
\begin{tabular}[c]{@{}l@{}}Multipack sampler \\ for data packing\end{tabular}          & \href{https://github.com/imoneoi/multipack_sampler}{https://github.com/imoneoi/multipack\_sampler}              \\ \midrule
BigVGAN \cite{lee2023bigvgan}                                  & \href{https://github.com/NVIDIA/BigVGAN}{https://github.com/NVIDIA/BigVGAN}              \\ \midrule
\begin{tabular}[c]{@{}l@{}}Metric Calculation \\ (WER, METEOR, ROUGE-L)\end{tabular} & \href{https://github.com/huggingface/evaluate}{https://github.com/huggingface/evaluate}              \\ \bottomrule
\end{tabular}
\end{center}
\end{table}

\subsubsection{Human Evaluation and GPT-4 Judge}
\label{appendix:mos}

As mentioned in Section \ref{experiments:evaluation}, we conduct various human evaluations and GPT-4-based assessments \cite{openai2023gpt4}. First, we employ Amazon Mechanical Turk to perform human preference tests and evaluate prosody and naturalness through P-MOS and MOS. In the human preference test, as detailed in the main paper, we present evaluators with previous spoken dialog and spoken input along with two candidate spoken responses. Evaluators are asked to choose which response is more appropriate, considering comprehensive aspects such as content, prosody, and sound quality. We provided evaluators with the instruction, \textit{``Given the spoken dialog of two speakers, which response is more suitable? Please consider comprehensive aspects such as content, speech quality, and prosody.''} All comparative experiments are evaluated by 150 evaluators, respectively, and the total cost for these evaluations is approximately $\$200$. 

We also conduct a qualitative evaluation by measuring the 5-scale mean opinion score (MOS) and prosody mean opinion score (P-MOS), both ranging from 1 to 5 points. For the P-MOS, we provide evaluators with the input spoken dialog and the corresponding ground truth text response. They are then asked to listen to the speech matching the ground truth text response and evaluate the prosody, considering both the spoken dialog and the response text. Additionally, we measure the MOS to judge audio quality and naturalness. In this scenario, evaluators are given only the response text and its corresponding spoken response without any preceding spoken dialog and are asked to rate the audio quality and naturalness. The instructions provided for the P-MOS and MOS tests are: \textit{``How natural is the prosody in this recording? Please focus on the prosody in the context of the spoken conversation flow and the given text response, and ignore other aspects such as speaker ID and sound quality.''} and \textit{``How natural (i.e., human-sounding) is this recording? Please focus on the audio quality and the naturalness of pronunciation.''}, respectively. 198 evaluators participate in the P-MOS measurement, and another 176 participate in the MOS measurement. For all 5-scale evaluations, we provide examples of speech rated as 1, 3, and 5 points as a reference to guide evaluators. We spend a total of approximately $\$250$ on these evaluations.

\begin{figure}
    \centering
    \includegraphics[width=0.99\linewidth]{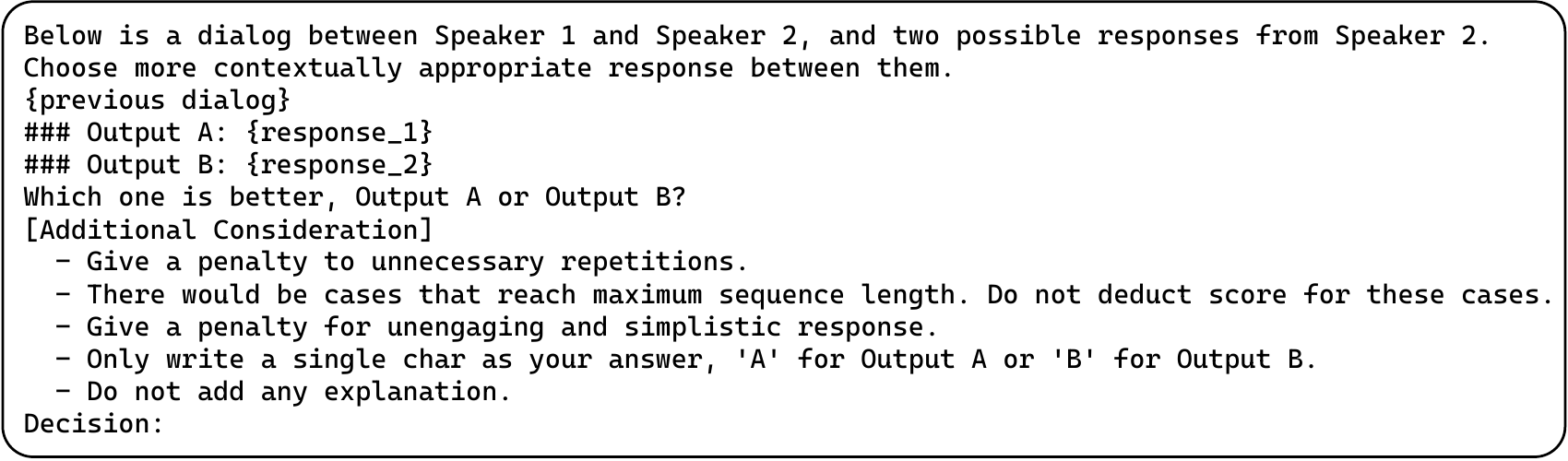} 
    \caption{Prompt used for LLM-based evaluation utilizing GPT-4.}
    \label{fig:gpt4_evaluation}
\end{figure}

To evaluate the semantic quality of the audio generated by each model, we utilize transcripts obtained by passing the generated audio through a separate ASR model, \textit{whisper-large-v3}. These transcripts are then evaluated using GPT-4 \cite{openai2023gpt4}. As introduced by \citet{zheng2023judging}, we provide the GPT-4 model with evaluation instructions, previous dialog, and two candidate responses each from USDM and a baseline model for comparison, asking it to choose the preferred response. The template used for this evaluation is shown in Figure \ref{fig:gpt4_evaluation}.

Among all available APIs, we use \textit{gpt-4-0125-preview} for evaluation. To avoid bias due to the order of response candidates, we assess the responses from the two models in both their original and reversed orders. Preference is primarily judged based on content appropriateness, but penalties are assigned for unengaging responses such as simple short answers and fillers. If the results differ between the two evaluations, it is marked as a Tie, if both prefer Output A, then A, and if both prefer Output B, then B, with results in Table \ref{table:model_comparion_quantitative}.

All $p$-values in the main text are obtained through the Wilcoxon signed-rank test. For the preference test, a score of 1 is assigned if a model is preferred, 0 for a tie, and -1 if not chosen. We then conduct a test using these scores and report the respective $p$-values.

\begin{table}[]
\caption{METEOR and ROUGE-L results measured using the text obtained from ASR of the spoken response (Transcribed Response) and results measured using the intermediate text response (Intermediate Response).}
\label{table:intermediate}
\small
\begin{center}
\begin{tabular}{l|ccc|cc}
\toprule
\multirow{2}{*}{\textbf{Method}} & \multicolumn{3}{c|}{\textbf{Transcribed Response}}    & \multicolumn{2}{c}{\textbf{Intermediate Response}} \\ \cmidrule{2-6} 
                                 & \textbf{METEOR} & \textbf{ROUGE-L} & \textbf{TTS WER} & \textbf{METEOR}         & \textbf{ROUGE-L}         \\ \midrule
Ground Truth                     & $-$             & $-$              & $2.2\%$          & $-$                     & $-$                      \\
USDM                             & 13.1            & 15.7             & $2.0\%$          & 13.8                    & 16.5                     \\
Cascaded                         & 12.5            & 15.0             & $1.3\%$          & 12.9                    & 15.5                     \\
From Scratch                     & 8.6             & 10.6             & $64.0\%$         & 10.6                    & 13.0                     \\
SpeechGPT                        & 9.9             & 11.8             & $23.2\%$         & 12.1                    & 13.8                     \\  \bottomrule
\end{tabular}
\end{center}
\end{table}

\subsection{Additional Experiments and Results}

\subsubsection{Additional Results for DailyTalk}
\label{appendix:intermediate}
In Section \ref{experiments:evaluation}, we assess the semantic quality by using the transcribed text of the spoken response generated by each model, using the \textit{whisper-large-v3}, and measure METEOR and ROUGE-L scores. Our models and baselines first generate a text response either within the end-to-end pipeline or through a separate model. We also measure METEOR and ROUGE-L scores for these intermediate text responses on the test sets of DailyTalk, and the results are presented in Table \ref{table:intermediate}.

Errors occur during the generation of the spoken response using intermediate text and the transcription of that spoken response for evaluation, leading to a performance gap between the results measured using the intermediate text response and the transcript of the spoken response. Despite the error gap, we confirm that USDM outperforms baselines in terms of the semantics of the intermediate response. Notably, a higher TTS WER increases the gap between the results based on the intermediate text response and the semantic performance of the final spoken response.

\subsubsection{Ablation Studies for Pretraining Scheme}
\label{appendix:ablations}

\begin{table}[]
\caption{Six types of speech-text interleaved sequences used to evaluate the performance of the pretrained model, along with the templates used for measuring PPL. For sequences with a continuation relationship, the speech and text data are split in half, combining one modality from the first half (e.g., \texttt{speech1 token} or \texttt{text1 token}) with the remaining modality from the second half (e.g., \texttt{text2 token} or \texttt{speech2 token}).}
\label{table:pretraining_eval_template}
\small
\begin{center}
\begin{tabular}{l|l}
\toprule
\multicolumn{1}{c|}{\textbf{Sequence}} & \multicolumn{1}{c}{\textbf{Template}} \\ \midrule
Unconditional Text                     &      \texttt{text token}                                 \\ \midrule
Unconditional Unit                     &       \texttt{speech token}                                \\ \midrule
Correspondence - Unit-to-Text          &      \texttt{speech token <|correspond|> text token}                                 \\ \midrule
Correspondence - Text-to-Unit          &     \texttt{text token <|correspond|> speech token}                                  \\ \midrule
Continuation - Unit-to-Text            &    \texttt{speech1 token <|continue|> text2 token}                                   \\ \midrule
Continuation - Text-to-Unit            &     \texttt{text1 token <|continue|> speech2 token}               \\ 
\midrule
\end{tabular}
\end{center}
\end{table}

\begin{table}
\caption{PPL of various pretraining schemes for diverse unit and text combinations for the \texttt{test-clean} subset of LibriSpeech. T2U represents text-to-unit, and U2T represents unit-to-text, with PPL measured only for the subsequent modality. Lower is better.}
\label{table:pretrained_ppl}
\small
\begin{center}
\begin{tabular}{l|cc|cc|cc|cc}
\toprule
\multirow{2}{*}{\textbf{Method}} & \multicolumn{2}{c|}{\textbf{Overall}} & \multicolumn{2}{c|}{\textbf{Unconditional}} & \multicolumn{2}{c|}{\textbf{Correspondence}} & \multicolumn{2}{c}{\textbf{Continuation}} \\ \cmidrule{2-9} 
                                 & \textbf{Text} & \textbf{Unit} & \textbf{Text}    & \textbf{Unit}    & \textbf{U2T}      & \textbf{T2U}     & \textbf{U2T}    & \textbf{T2U}    \\ \midrule
Ours                             & 6.886             & 4.813             & 17.175               & 5.037                & 1.133                 & 4.113                & 16.781              & 5.380               \\
Setup 1                & 14.485            & 5.261             & 17.195               & 5.047                & 11.578                & 5.345                & 15.267              & 5.398               \\
Setup 2                  & 31.679            & 5.619             & 17.846               & 5.107                & 1.108                 & 4.098                & 1607.743            & 6.600               \\
Setup 3                   & 21.392            & 5.146             & 17.463               & 5.086                & 1.107                 & 4.110                & 506.374             & 6.521       \\ 
\bottomrule
\end{tabular}
\end{center}
\end{table}

In this section, we provide further explanation of the ablation studies on the pretraining schemes discussed in Section \ref{experiments:pretraining}. We design interleaved sequences excluding each key relationship, continuation and correspondence, to demonstrate the necessity of each relationship within our proposed speech-text pretraining scheme. Additionally, we follow the cross-modal pretraining scheme proposed in Spectron \cite{nachmani2023spoken}, which we name Setup 3.

\textbf{Setup 1} We create interleaved speech-text sequences composed solely of continuation relationships and use these for speech-text pretraining. The interleaved sequences used for Setup 1 can be obtained by skipping the last step of the 3-step data preparation process described in Section \ref{approach:unified-speech-text-pretraining}. This approach is similar to previous works such as BLSP \cite{wang2023blsp} and SpiRit-LM \cite{nguyen2024spiritlm}.

\textbf{Setup 2} We construct cross-modal sequences exclusively with correspondence relationships. The interleaved sequences with this relationship are typically formatted as \texttt{``speech token <|correspond|> text token''} and \texttt{``text token <|correspond|> speech token''} sequences, similar to SpeechGPT \cite{zhang2023speechgpt}.

\textbf{Setup 3}  We also compare a scheme that utilizes one fixed template for pretraining. Following Spectron \cite{nachmani2023spoken} we pretrain using an interleaved sequence where the input speech is transcribed into text, followed by predicting the subsequent response text and synthesizing the corresponding speech. Assuming the speech and the corresponding text are split into two parts (speech1, text1, speech2, text2), we perform cross-modal pretraining using the interleaved sequence \texttt{``speech1 token <|correspond|> text token <|correspond|> speech2 token''}, where the \texttt{text token} is obtained by concatenating text1 and text2.

Each model is trained with the same data and hyperparameters as our pretraining. All models have the same vocabulary size. We measure the PPL of various combinations of speech-text sequences created using the LibriSpeech \texttt{test-clean} subset. We create 6 types of interleaved sequences and the templates of these sequences are listed in Table \ref{table:pretraining_eval_template}. For measuring the PPL of text tokens, we normalize the probability by excluding the probability of units and use this normalized probability to measure the PPL. Similarly, for speech modality PPL, we compute the logits and probabilities for only the unit tokens, which have a vocabulary size of 10,000, and then calculate the PPL. To evaluate pretraining schemes where only one of the special tokens \texttt{<|correspond|>} or \texttt{<|continue|>} is used, we insert the special token used during pretraining at the boundary between the two modalities, regardless of the combination being evaluated.

The PPL for each combination is listed in Table \ref{table:pretraining_eval_template}. Setup 1 and Setup 2, which model only one relationship, fail to model the other and exhibit high PPL values. Additionally, Setup 3, which uses a specific fixed template of interleaved sequences for pretraining, shows superior performance in interleaved sequences with a correspondence relationship but is unable to model a continuation relationship. In contrast, our speech-text model, which universally models various relationships, demonstrates consistently powerful performance regardless of the sequence type.

\subsubsection{Per-Task Training Dynamic Analysis}
\label{appendix:analysis}

\begin{figure}
    \centering
    \includegraphics[width=0.41\linewidth]{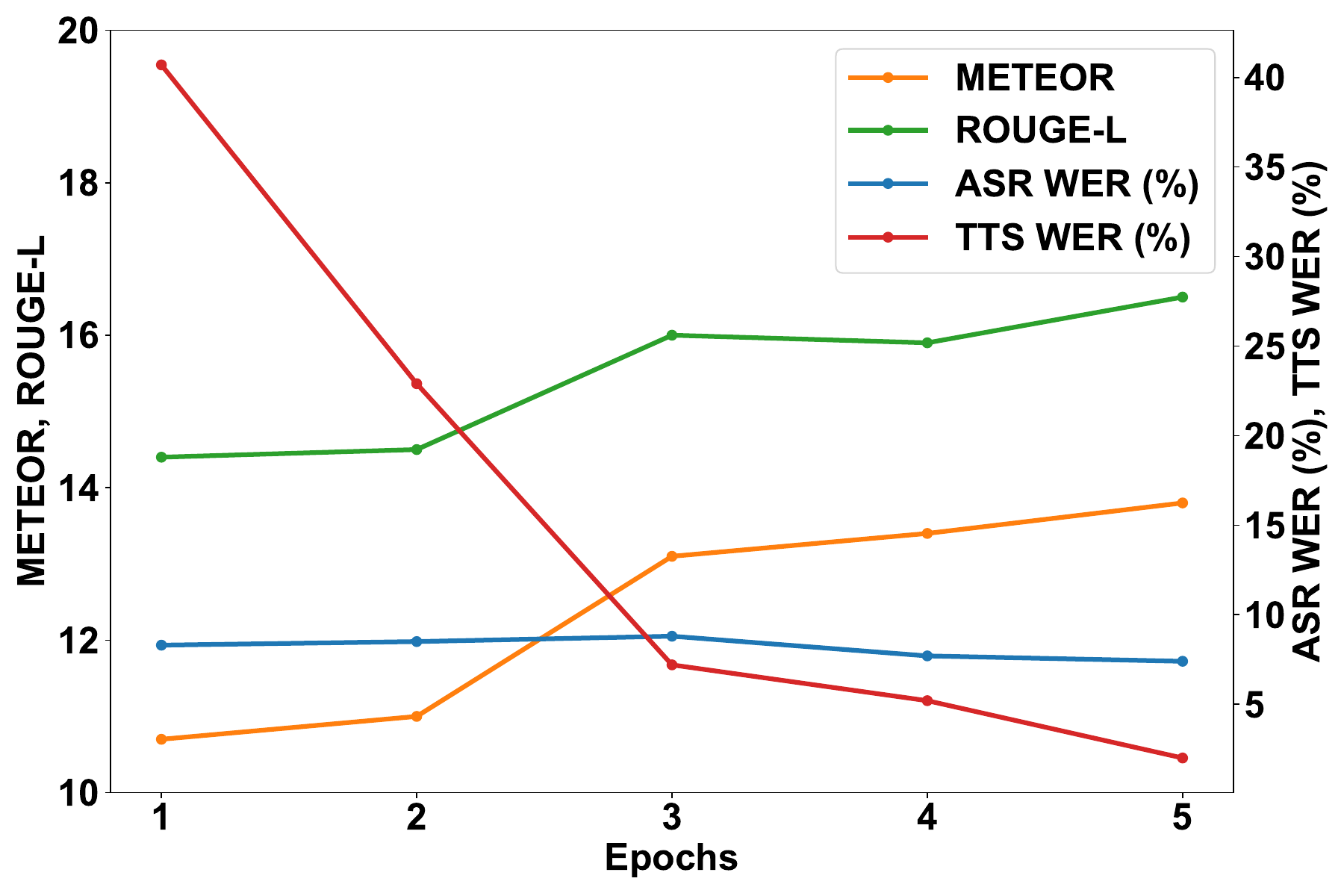} 
    \hspace{0.1cm}
    \includegraphics[width=0.47\linewidth]{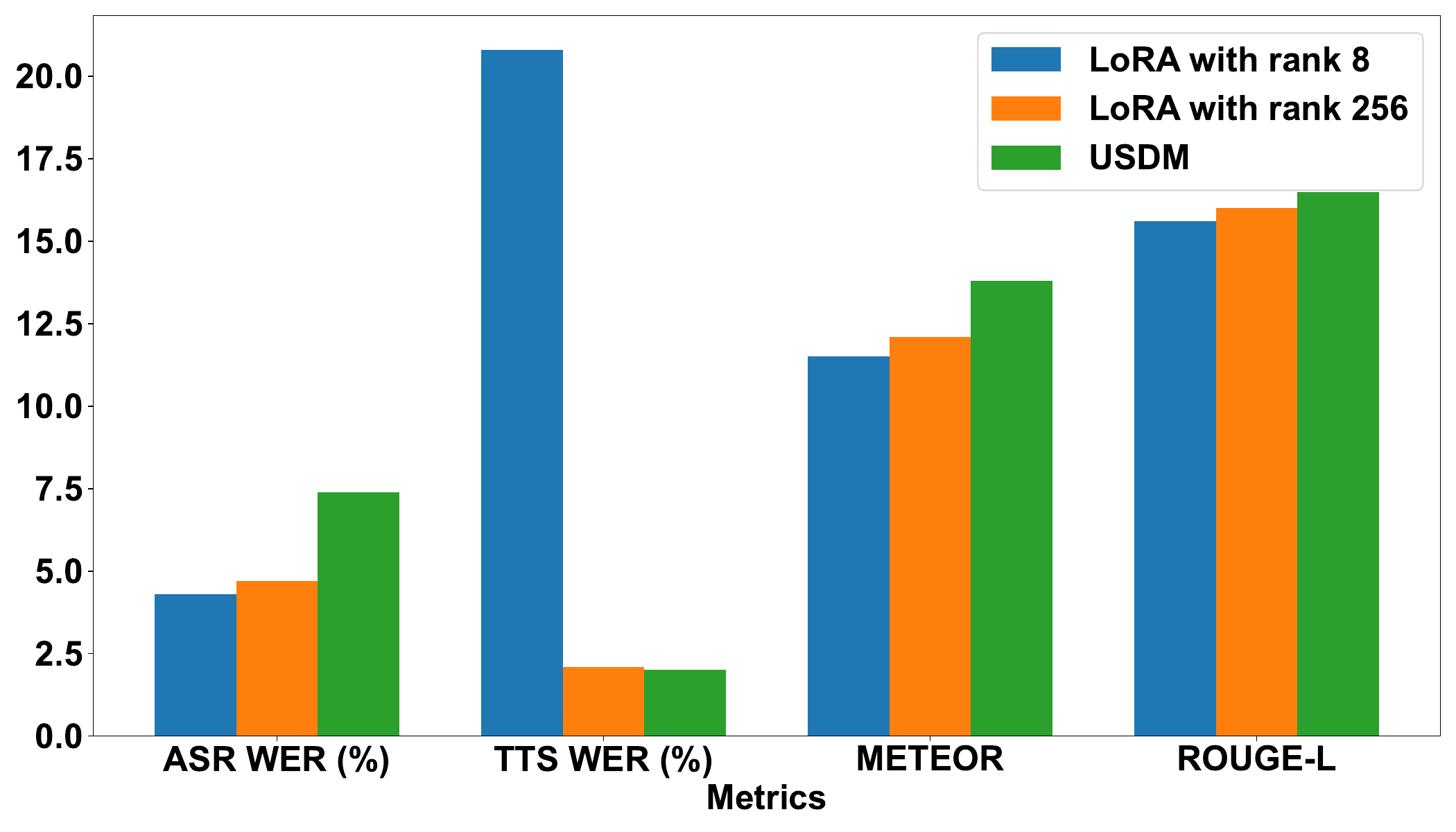} 
    \caption{Left is the quantitative results for each epoch of the USDM fine-tuned on DailyTalk. The figure on the right illustrates the performance of the Spoken Dialog Model when trained with Low-Rank Adaptation (LoRA) versus full fine-tuning.}
    \label{fig:dynamics}
    \vskip -0.2in
\end{figure}

Our model adopts an end-to-end pipeline with intermediate text, where the input speech is first transcribed, followed by generating the response text. Consequently, the model simultaneously learns unit-to-text, text response generation, and text-to-unit, with each task potentially reaching its optimal point at different epochs. To observe the training dynamics of each task, we train the USDM on the DailyTalk dataset for 5 epochs and monitor TTS WER, STT WER, METEOR, and ROUGE-L at each epoch. As shown on the left side of Figure \ref{fig:dynamics}, while STT WER remains consistent over each epoch, TTS WER, METEOR, and ROUGE-L scores improve, suggesting that the dialog modeling task and text-to-unit tasks are more challenging compared to the unit-to-text task.

Additionally, we build two spoken dialog models using Low-Rank Adaptor (LoRA) \cite{hu2022lora} for fine-tuning the pretrained speech-text model on the DailyTalk dataset, and observe similar tendencies. We fine-tune the model using a higher learning rate of $10^{-4}$, comparable to USDM, with LoRA ranks of 8 and 256. Consistently, as the right side of Figure \ref{fig:dynamics} illustrates, increasing the number of fine-tuning parameters improves semantic performance in dialog and unit synthesis performance but deteriorates the performance in the unit-to-text task. Considering these observations and the analysis in Section \ref{experiments:analysis} that demonstrates the importance of unit-to-text performance in USDM, we plan to explore strategies to mitigate overfitting in the unit-to-text task by varying the loss weight for each task within the pipeline or by applying the curriculum learning approach in future work.

\begin{figure}
    \centering
    \includegraphics[width=0.99\linewidth]{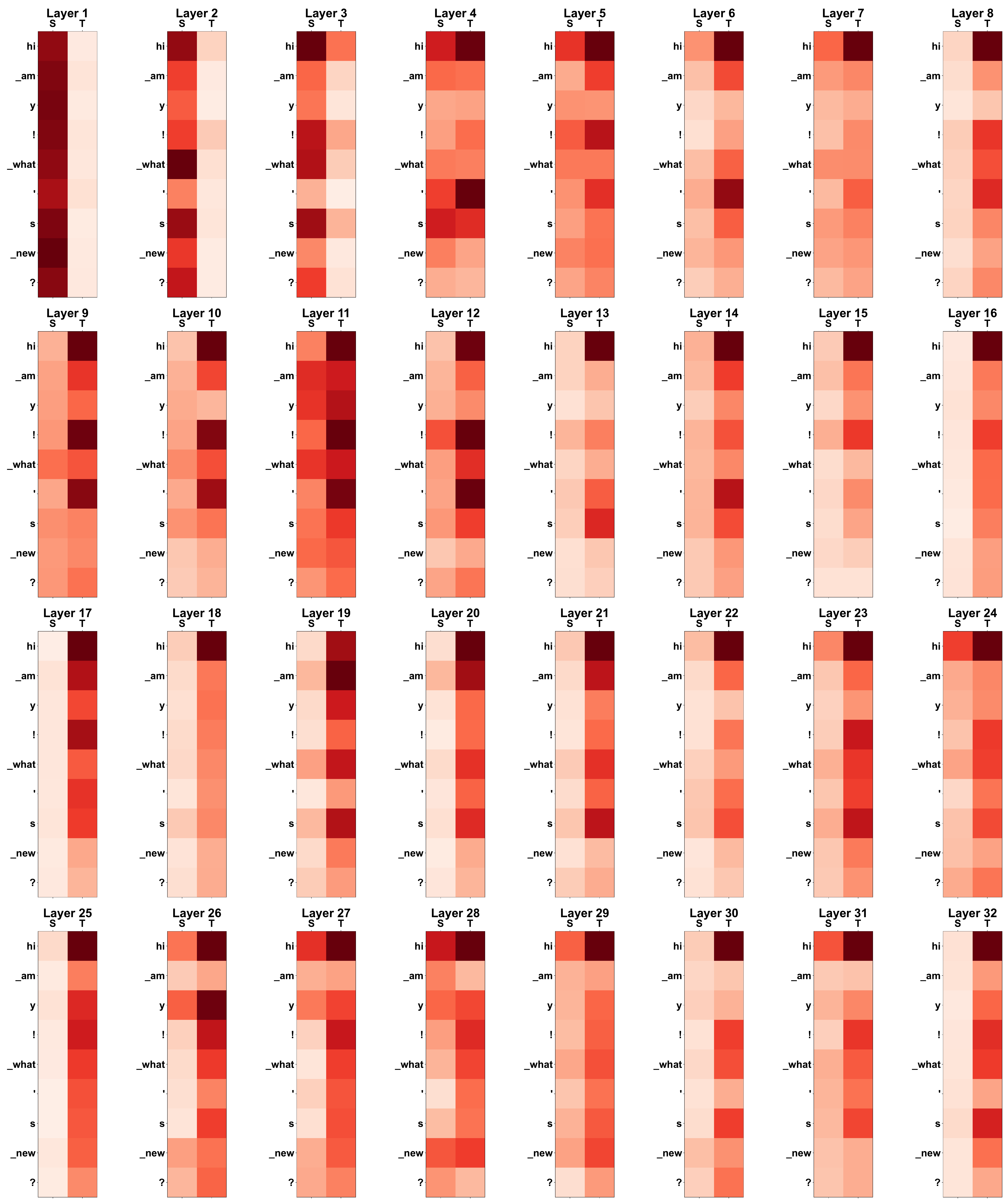} 
    \caption{Attention map plots for the USDM response to the input speech ``Hi George! It's good to see you!''. We plot attention maps for all layers as described in Section \ref{experiments:analysis}. Although there are variations in intensity, we observe in all layers that the response text attends to both the speech input and the transcribed text.}
    \label{fig:attention_map}
\end{figure}

\subsubsection{Additional Attention Maps}
We plot the attention maps of the generated text response to the speech input and its model-generated transcript for 6 selected layers in Figure \ref{fig3}. Given that our spoken dialog model consists of 32 layers, we additionally include attention maps of another sample for all layers in Figure \ref{fig:attention_map}.


\end{document}